\documentclass[journal]{IEEEtran}
\usepackage{ifpdf}

\usepackage{cite}

\ifCLASSINFOpdf
  \usepackage[pdftex]{graphicx}
  \graphicspath{{./figures/}}
  \DeclareGraphicsExtensions{.pdf,.jpeg,.png,.jpg}
\else
\fi
\usepackage{amsmath}
\usepackage{algorithmic}

\usepackage{array}

\ifCLASSOPTIONcompsoc
  \usepackage[caption=false,font=normalsize,labelfont=sf,textfont=sf]{subfig}
\else
  \usepackage[caption=false,font=footnotesize]{subfig}
\fi

\usepackage{stfloats}
\ifCLASSOPTIONcaptionsoff
  \usepackage[nomarkers]{endfloat}
 \let\MYoriglatexcaption\caption
 \renewcommand{\caption}[2][\relax]{\MYoriglatexcaption[#2]{#2}}
\fi
\usepackage{url}

\newcommand{\norm}[1]{\left\lVert#1\right\rVert}
\newcommand{\abs}[1]{\left\lvert#1\right\rvert}

\newcommand{\R}[1]{\mathbb{R}^{#1}}
\newcommand{\ihv}{I^{\mathrm{HV}}}

\newcommand{\znadir}{z^{\mathrm{nad}}}
\newcommand{\zideal}{z^{\mathrm{ide}}}
\newcommand{\ntarget}[1]{N_{\tau^{#1}}}

\usepackage{amsmath}
\usepackage{amsfonts}

\usepackage{lipsum}

\usepackage[dvipsnames]{xcolor}

\def\revisions{ON}
\usepackage{ulem}
\def\tmprev{ON}
\newcommand{\ins}[1]{%
\ifx\revisions\tmprev
{\color{blue} #1}%
\else #1%
\fi%
}
\newcommand{\del}[1]{%
\ifx\revisions\tmprev
{\color{red}\sout{#1}}%
\fi%
}

\begin{document}
%
\title{Characterization of Constrained Continuous Multiobjective Optimization Problems:\\ A Performance Space Perspective}
%
%
%


\author{Aljoša~Vodopija, Tea~Tušar, and~Bogdan~Filipič
\thanks{The authors are with the Department
of Intelligent Systems, Jožef Stefan Institute, SI-1000 Ljubljana, Slovenia, and also with Jožef Stefan International
Postgraduate School, SI-1000 Ljubljana, Slovenia (e-mail: aljosa.vodopija@ijs.si, tea.tusar@ijs.si, bogdan.filipic@ijs.si).}
\thanks{The authors acknowledge the financial support from the Slovenian Research Agency (young researchers program and research core funding no.\ P2-0209). The authors also acknowledge the project ``Constrained Multiobjective Optimization Based on Problem Landscape Analysis'' (no.\ N2-0254) was financially supported by the Slovenian Research Agency.}}

%
%

\markboth{}%
{Vodopija \MakeLowercase{\textit{et al.}}: Characterization of CMOPs: A Performance Space Perspective}
%



\maketitle

\begin{abstract}
Constrained multiobjective optimization has gained much interest in the past few years. However, constrained multiobjective optimization problems (CMOPs) are still unsatisfactorily understood. Consequently, the choice of adequate CMOPs for benchmarking is difficult and lacks a formal background. This paper addresses this issue by exploring CMOPs from a performance space perspective. First, it presents a novel performance assessment approach designed explicitly for constrained multiobjective optimization. This methodology offers a first attempt to simultaneously measure the performance in approximating the Pareto front and constraint satisfaction. Secondly, it proposes an approach to measure the capability of the given optimization problem to differentiate among algorithm performances. Finally, this approach is used to contrast eight frequently used artificial test suites of CMOPs. The experimental results reveal which suites are more efficient in discerning between three well-known multiobjective optimization algorithms. Benchmark designers can use these results to select the most appropriate CMOPs for their needs.
\end{abstract}


%
\IEEEpeerreviewmaketitle

\section{Introduction} 
\label{sec:introduction}
\normalem 

\IEEEPARstart{M}{any} real-world continuous optimization problems involve the optimization of multiple, often conflicting objectives, and constraints that need to be respected~\cite{Ma2019}. Such problems are known as \emph{constrained multiobjective optimization problems} (CMOPs) and have recently gained much interest in the evolutionary computation community. Indeed, several novel techniques for constraint handling and new test suites of CMOPs have been proposed recently (e.g.,~\cite{Fan2019b, Zhu2020, Ma2021a, Zhi-Zhong2021}).

Despite the large amount of recently published articles in the field of constrained multiobjective optimization, the CMOPs for benchmarking \emph{multiobjective evolutionary algorithms (MOEAs)} and corresponding \emph{constraint handling techniques (CHTs)} are still unsatisfactorily understood and characterized~\cite{Picard2021, Vodopija22}. Consequently, the selection of appropriate CMOPs for benchmarking is difficult and lacks a formal background. In the circumstances, preparing a sound and well-designed experimental setup for constrained multiobjective optimization is a challenging task. A poorly designed benchmark might lead to inadequate conclusions about CMOP landscapes and MOEA performance~\cite{Vodopija22}. 

According to~\cite{Bartz-Beielstein2020}, there are two main options for characterizing and evaluating the quality of optimization problems, namely through the \emph{feature space} and \emph{performance space}. 
The feature space can be seen as a space of problem characteristics, including basic characteristics such as problem dimensionality and type of objective and constraint functions, as well as more advanced problem characteristics derived using methods developed in the field of \emph{exploratory landscape analysis (ELA)}~\cite{Mersmann2011}. On the other hand, the performance space represents the problems based on the obtained algorithm performance (behavior) while solving these problems. Similar to the feature space, basic statistics can be used, such as mean or median algorithm performance, as well as more advanced methods, e.g., \emph{data profiles}~\cite{More2009} or \emph{empirical cumulative distribution functions (ECDFs)}~\cite{Hansen21, Hansen2022}. In contrast to the aggregated values (means, medians, etc.), the latter two methods consider the progress of the whole algorithm run and, this way, provide more comprehensive information about the algorithm behavior. In our previous work~\cite{Vodopija22}, we provided an extensive study of characterizing CMOPs through the feature space, while, to the best of our knowledge, the performance space has not been  addressed in sufficient depth yet.

In the literature, the performance indicators used in constrained multiobjective optimization are the same as those used in unconstrained multiobjective optimization, they are simply applied only to feasible solutions. The most frequently employed indicators are the \emph{hypervolume indicator}~\cite{Zitzler1999} and \emph{inverted generational distance}~\cite{Bosman2003} since they can provide information about the convergence and the diversity of the obtained Pareto front approximation. For monitoring the performance during the run, one can use convergence graphs, data profiles or ECDFs. However, none of these techniques provides relevant information until feasible solutions are discovered. As a result, essential insights about the algorithm behavior and CMOP characteristics are missed. To overcome this situation, some papers report also the constraint satisfaction progress~\cite{Fan2019a}. Nevertheless, to the best of our knowledge, no method from the literature simultaneously measures both the convergence towards the Pareto front and constraint satisfaction, making the experimental analysis incomplete. 
 
Moreover, we are aware of only a single work analyzing the CMOPs from the performance space perspective which was conducted in 2017~\cite{Tanabe2017}. The authors used five CHTs to characterize five artificial and seven real-world test problems. The results revealed that only a single artificial test problem was suitable for benchmarking algorithms since the other four problems could be solved even without employing a CHT. Additionally, the studied real-world problems were inadequate since they could not differentiate among MOEAs—a desired property of a test problem as it provides relevant information for algorithm designers~\cite{Bartz-Beielstein2020}. Since 2017, several novel test suites of CMOPs have been proposed, and their ability to differentiate among MOEAs has not been investigated yet~\cite{Filipic21}. 

In this paper, we present a novel anytime performance assessment approach specifically designed for constrained multiobjective optimization. It simultaneously monitors both the Pareto front approximation and constraint satisfaction. The approach is inspired by the anytime performance assessment of algorithms on unconstrained bi-objective optimization problems used in COCO (COmparing Continuous Optimizers)~\cite{Hansen21}. In addition, we propose an approach to measure the capability of a given problem to differentiate among MOEAs. The resulting measure is then used to evaluate the most frequently used artificial test suites of CMOPs. Because of space limitations, we present only selected results in this paper. The interested reader can find the complete results online\footnote{https://vodopijaaljosa.github.io/cmop-web/}.

The rest of this paper is organized as follows. In Section~\ref{sec:theoretical_background}, we provide the theoretical background for constrained multiobjective optimization and introduce the COCO platform Then, in Section~\ref{sec:methodology}, we extend the performance assessment from the COCO platform to CMOPs and propose an approach to characterize CMOPs based on the algorithm performance. Section~\ref{sec:experiments} provides details on the experimental setup, while Section~\ref{sec:results} presents the results, evaluates the existing test suites of CMOPs, and discusses some limitations of the proposed methodology. Finally, a summary of findings and ideas for future work are discussed in Section~\ref{sec:conclusions}.

\section{Background} 
\label{sec:theoretical_background}

In this section, we provide the theoretical background for this work. After the definitions for CMOPs and constraint violation, we describe the performance assessment approach for multiobjective optimization used in the COCO platform.

\subsection{Constrained Multiobjective Optimization Problems}

A CMOP is, without loss of generality, formulated as:
\begin{equation} \label{eq:cmop}
\begin{split}
    \text{minimize} \quad &f_m(x), \quad m = 1, \dots, M \\
    \text{subject to} \quad &g_i(x) \leq 0, \quad i = 1, \dots, I\\
\end{split}
\end{equation}
where $x = (x_1, \dots, x_D)$ is a \emph{search vector}, $f_m: S \rightarrow \R{}$ are \emph{objective functions}, $g_i: S \rightarrow \R{}$ \emph{constraint functions}, $S \subseteq \R{D}$ is a \emph{search space} of dimension $D$, and $M$ and $I$ are the numbers of objectives and constraints, respectively. 

In particular, when $M=1$ the corresponding problem is a \emph{constrained single-objective optimization problem (CSOP)}. To differentiate between CMOPs and CSOPs, in the later case, we will omit the index $m$ ($f = f_1$). Additionally, if the problem has no constraints, it is called a \emph{single-objective optimization problem (SOP)} when $M = 1$, and a \emph{multiobjective optimization problem (MOP)} when $M > 1$.

One of the most important concepts in constrained optimization is the notion of the \emph{constraint violation}. For a single constraint $g_i$ it is defined as $v_i(x) = \max(0, g_i(x))$ and combined for all constraints together as
\begin{equation}
    v(x) = \sum_{i=1}^{I} v_i(x)
\end{equation}
into the \emph{overall constraint violation}. A solution $x$ is feasible iff its overall constraint violation equals zero ($v(x) = 0$).

Note that other definitions for overall constraint violation exist, and their use would impact the analysis performed in this study. However, the proposed definition for the overall constraint violation is by far the most commonly adopted in constrained optimization~\cite{Filipic21}, and as such, it represents the most appropriate choice.

A feasible solution $x \in S$ \emph{dominates} another solution $y \in S$ iff $f_m(x) \leq f_m(y)$ for all $1 \leq m \leq M$ and $f_m(x) < f_m(y)$ for at least one $1 \leq m \leq M$. Additionally, a solution $x^* \in S$ is \emph{Pareto optimal} if there is no solution $x \in S$ such that dominates $x^*$. We can generalize the Pareto dominance to sets. A set $X$ dominates set $Y$ iff for each $y \in Y$ there exists at least one solution $x \in X$ that dominates $y$. 

The set of all feasible solutions is called the \emph{feasible region} and is denoted by $F = \{x \in S \mid v(x) = 0\}$. All nondominated feasible solutions represent a \emph{Pareto-optimal set}, $S_\text{o}$. The image of the Pareto-optimal set in the objective space is the \emph{Pareto front} and is denoted here by $P_\text{o} = \{f(x) \mid x \in S_\text{o}\}$.

The \emph{ideal objective vector}, $\zideal$, is defined as the vector in the objective space that contains the optimal objective value for each objective separately and it is expressed as 
\begin{equation}
    \zideal = \left(\inf_{x \in F}f_1(x), \dots, \inf_{x \in F}f_M(x)\right).
\end{equation}
Additionally, the \emph{nadir objective vector}, $\znadir$, consists in each objective of the worst value obtained by any Pareto-optimal solution. It can be expressed as
\begin{equation}
    \znadir = \left(\sup_{x \in P_\text{o}}f_1(x), \dots, \sup_{x \in P_\text{o}}f_M(x)\right).
\end{equation}
An additional important concept is the region of interest in the objective space, $Z$, which represents the set of objective vectors bounded by the ideal and nadir objective vectors.

If good approximations for ideal and nadir objective vectors are known, the objective functions can be normalized to
\begin{equation}
    \frac{f_m(x) - \zideal_m}{\znadir_m - \zideal_m}.
\end{equation}
This way, the objective values are of approximately the same magnitude and the range of the objective values for Pareto-optimal solutions is $[0,1]$. Note that after normalization the $\zideal$ consists of $m$ zeros ($\zideal = (0, \dots, 0)$), $\znadir$ of $m$ ones ($\znadir = (1, \dots, 1)$), and the region of interest $Z$ equals $[0,1]^M$. In particular, for SOPs and CSOPs the normalization results in $f(x^*) = 0$. In the rest of this paper, we assume that all the objective functions are normalized.

\subsection{Empirical Runtime Distributions (ERDs)}
\label{sec:coco}

The performance measurement approach used in the COCO framework~\cite{Hansen21, Hansen2022} relies on the number of function evaluations\footnote{When we refer to a function evaluation, we actually mean the evaluation of all the objective and constraint functions. For example, for a bi-objective problem with three constraints we need to perform five evaluations, however, we count this as a single function evaluation.}---called \emph{runtime}---needed for an algorithm, $a$, to reach predefined quality indicator targets. More precisely, we can present an algorithm run after preforming $T$ function evaluations as a sequence of candidate solutions, $A^T(a) = \{x^1(a), \dots,  x^T(a)\}$. Within this framework, a quality indicator, $I$, is defined as a function mapping  $A^T(a)$ to a real value. Here, we assume that low quality indicator values indicate better sequences of candidate solutions, and vice versa. Additionally, the runtime for a given quality indicator target equals to the lowest $T$ for which $I(A^T(a))$ reaches the given target precision value, $\tau$. Note that in the following, if there is no ambiguity, we remove the algorithm notation $a$ from $A^T(a)$.

In practice, we define several target precision values to understand the algorithm behavior through the entire run. Runtimes can be formally defined as random variables and displayed as an empirical cumulative distribution function---called empirical runtime distribution (ERD) in the COCO framework. ERDs are used to display the proportion of target values reached within a specified budget and can be easily aggregated over multiple restarts, runs or even multiple problems. For more details on ERDs, see~\cite{Hansen21, Hansen2022}. The runtime data set for an algorithm $a$ and all targets $\tau$-s is denoted as $\{T_a(\tau)\}_{\tau}$. Finally, the runtimes in COCO are usually studied in a logarithmic scale and this perspective is used throughout this paper as well.

\subsection{Quality Indicators}

Based on the nature of the optimization problem there are various quality indicators used by the COCO framework. Those relevant for this work are as follows.

\subsubsection{Single-objective optimization} 
In this case, the quality indicator is the best so far observed objective function value:
\begin{equation} \label{eq:isop}
    I^\mathrm{SOP}(A^T) = \min_{x \in A^T} f(x).
\end{equation}

\subsubsection{Constrained single-objective optimization}
The quality indicator for unconstrained problems (\ref{eq:isop}) is extended by the addition of the overall constraint violation as follows:
\begin{equation} \label{eq:icsop}
    I^\mathrm{CSOP}(A^T) = \min_{x \in A^T} f(x) + v(x).
\end{equation}

\subsubsection{Multiobjective optimization}
\begin{figure*}[!t]
    \centering
    \subfloat[]{\includegraphics[width=0.31\textwidth]{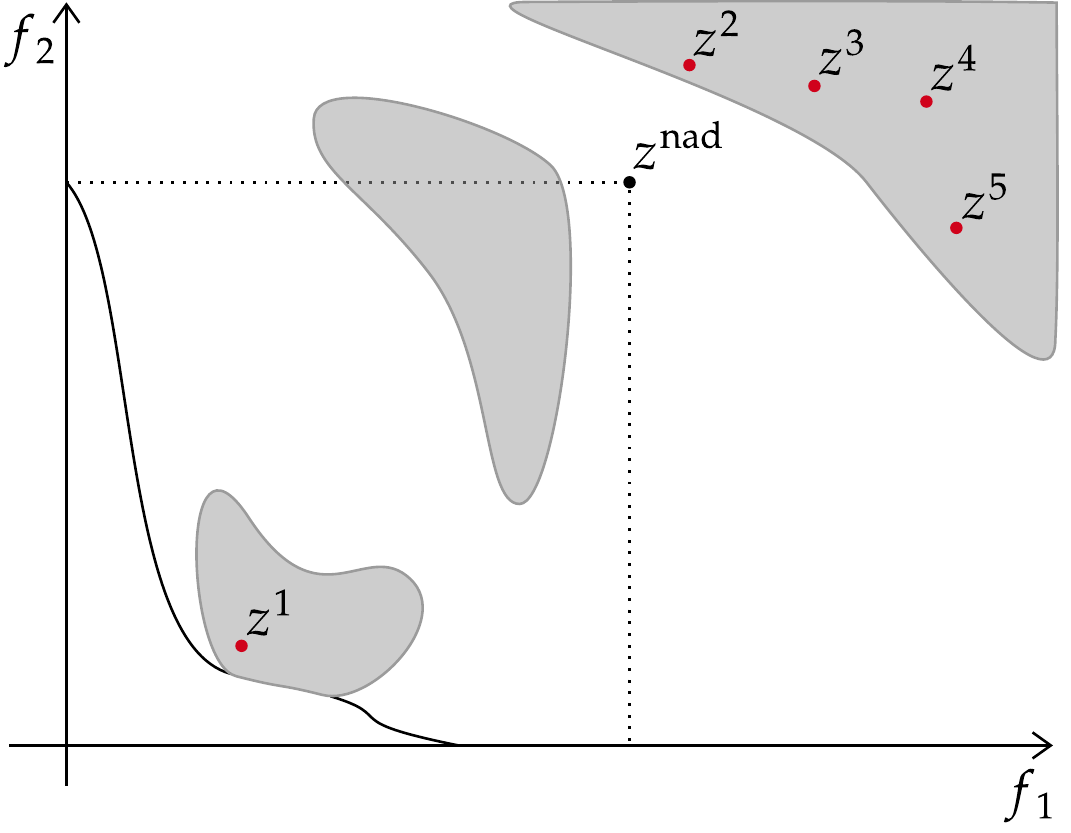}%
    \label{fig:1a}}
    \hspace{0.025\textwidth}
    \subfloat[]{\includegraphics[width=0.31\textwidth]{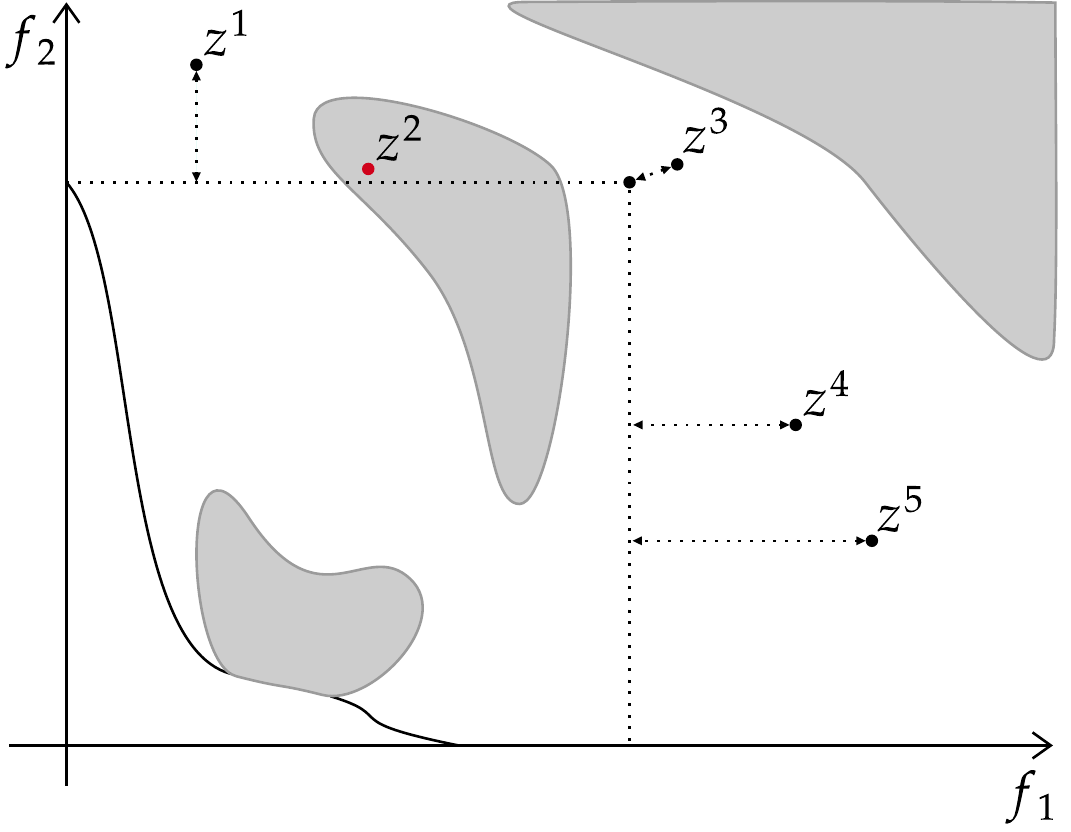}%
    \label{fig:1b}}
    \hspace{0.025\textwidth}
    \subfloat[]{\includegraphics[width=0.31\textwidth]{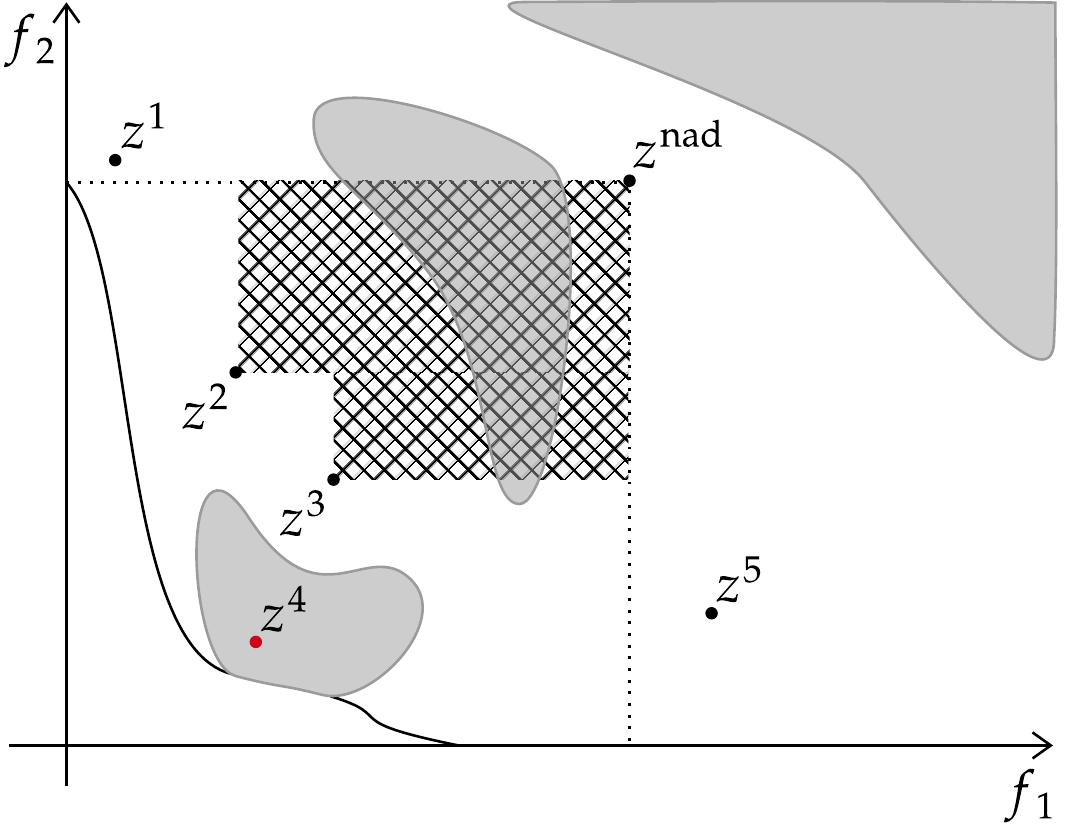}%
    \label{fig:1c}}
    \caption{The quality indicator $I^\mathrm{CMOP}$ at three stages of the algorithm search: (a) All the solutions belong to the infeasible space (areas in gray) and the quality indicator relies on the overall constraint violation. (b) There exists at least one feasible solution but no solutions dominate the reference point $\znadir$. The quality indicator relies on the distance to the region of interest $Z$ (area bounded by the doted lines and the coordinate axes). (c) There exists at least one feasible solution dominating the reference point. The quality indicator is based on the hypervolume (area depicted with a mesh).}
    \label{fig:1}
\end{figure*}

The quality indicator for MOPs consists of two parts. When no solution from the sequence $A^T$ dominates the nadir point (reference point), the distance to the region of interest $Z$ is used to measure the quality of the solutions (see Fig.~\ref{fig:1b}). In contrast, when at least one of the solutions dominates the nadir point, the hypervolume indicator is used instead (see Fig.~\ref{fig:1c}). This quality indicator can be mathematically expressed as: 
\begin{equation} \label{eq:imop}
    I^\mathrm{MOP}(A^T) = 
    \begin{cases}
        - \ihv(A^T) & \text{if } A^T \preceq \{ \znadir \} \\
        d(A^T, Z) & \text{otherwise}.
    \end{cases}
\end{equation}
Here,
\begin{equation}
    \ihv(A^T) = V \left(\bigcup_{x \in \mathcal{N}(A^T)} [f_1(x), 1] \times \cdots \times [f_M(x), 1] \right)
\end{equation}
represents the hypervolume of the archive $A^T$, $\mathcal{N}(A^T)$ is the set of all the points from $A^T$ dominating the reference point which equals $(1, \dots, 1)$, and
\begin{equation} \label{eq:dist}
    d(A^T, Z) = \inf_{(x,z) \in A^T \times Z} \norm{f(x) - z}
\end{equation}
is the smallest Euclidean distance between the archive and the region of interest $Z$. Additional information on this quality indicator can be found in~\cite{Hansen2022}. 

\subsection{Target Precision Values for MOPs}
For each problem a set of quality indicator target values is chosen, which is used to measure algorithm runtimes and in turn to calculate ERDs. The target values are computed in the form of $\tau(\varepsilon) = \tau^\mathrm{ref} + \varepsilon$, where $\tau^\mathrm{ref}$ is a reference $I^\mathrm{MOP}$ value. It is either based on the hypervolume of the true Pareto front or an estimation for it. In COCO, 51 positive target precision values $\varepsilon \in \{10^{-5}, 10^{-4.9}, \dots, 10^{-0.1}, 10^{0}\}$ are chosen\footnote{In COCO, negative values are also introduced in case the algorithms find better Pareto front approximations than available ones. However, in our situation this cannot happen as all the Pareto fronts are known in advance.}. Note that it is not uncommon that the quality indicator value of the algorithm never reaches some of these target values, which leads to missing runtime measurements.

\section{Methodology} 
\label{sec:methodology}

This section provides an extension of ERDs to constrained multiobjective optimization. It also discusses an approach to measure the problem's effectiveness to distinguish algorithms based on the distance between ERDs.

\subsection{Quality Indicator for CMOPs}
There are two main paradigms to approach constrained optimization problems. The first one is applicable when the constraints must be satisfied at any cost, while the second one allows for partial violation of constraints if the objective values of a solution are of good quality. Although both paradigms have their pros and cons, we study the former one as this is the prevalent approach in the literature\footnote{The COCO framework uses the second paradigm for CSOPs.}.  

Furthermore, in many real-world scenarios the objective values cannot be calculated if the solution is infeasible~\cite{Eiben2003}. This often happens in simulation-based optimization where the simulator cannot return meaningful results if some of the constraints are not satisfied. Consequently, our main assumption in constructing a quality indicator for constrained multiobjective optimization is that an infeasible solution is strictly worse than any feasible solutions regardless of the objective value quality---this is exactly the pillar of the first paradigm. For example, in Fig.~\ref{fig:1c}, solution $z^4$ has better objective values than $z^5$. Actually, if no constraints were considered, $z^4$ would dominate $z^5$. Nevertheless, $z^5$ is considered to be strictly better than $z^4$. This desired property of a quality indicator can be mathematically expressed as follows:
\begin{equation} \label{eq:ic}
    I(x) < I(y) \quad \text{ for all } (x,y) \in F \times S \backslash F.
\end{equation}

The quality indicator for CSOPs (\ref{eq:icsop}) does not satisfy this property. The biggest disadvantage of an indicator not satisfying (\ref{eq:ic}) is that no matter how small the quality indicator value is, we cannot know whether there exists a feasible solution in $A^T$. For example, for a certain CSOP there might exist a solution in $A^T$ with $f(x) = 0$ and an arbitrarily small overall constraint violation value. In other words, unless $I^\mathrm{CSOP}$ equals zero ($x^* \in A^T$), we cannot know for certain whether we found a feasible solution relying solely on the quality indicator values. From a practical point of view, we wish for a quality indicator to involve a threshold that unequivocally indicates when the algorithm reached the feasible space.

Considering this, we propose an extension of the quality indicator for MOPs (\ref{eq:imop}) as follows:
\begin{equation} \label{eq:icmop}
    I^\mathrm{CMOP}(A^T) = 
    \begin{cases}
        \min(I^\mathrm{MOP}(A^T \cap F), \tau^*)  & A^T \cap F \neq \emptyset\\
        \min_{x \in A^T} v(x) + \tau^* & \text{otherwise} 
    \end{cases}
\end{equation}
where $\tau^*$ is a threshold to indicate that the feasible space was reached. For example, in the COCO framework, it can be set to the largest considered target for MOPs, which equals 1. It is also obvious that $I^\mathrm{CMOP}$ satisfies the property (\ref{eq:ic}). The behavior of the proposed quality indicator is illustrated in Fig.~\ref{fig:1}.

Additionally, note that in (\ref{eq:icmop}) only feasible solutions are considered in the calculation of $I^\mathrm{MOP}$. This can be seen in Figs.~\ref{fig:1b}~and~\ref{fig:1c} where the infeasible solutions are not considered for the calculation of the distance and hypervolume, respectively, once feasible solutions have been found. 

\subsection{Performance Space Comparison}
\label{sec:performance_space_comparison}
\begin{figure}[!t]
    \centering
    \includegraphics[width=0.62\columnwidth]{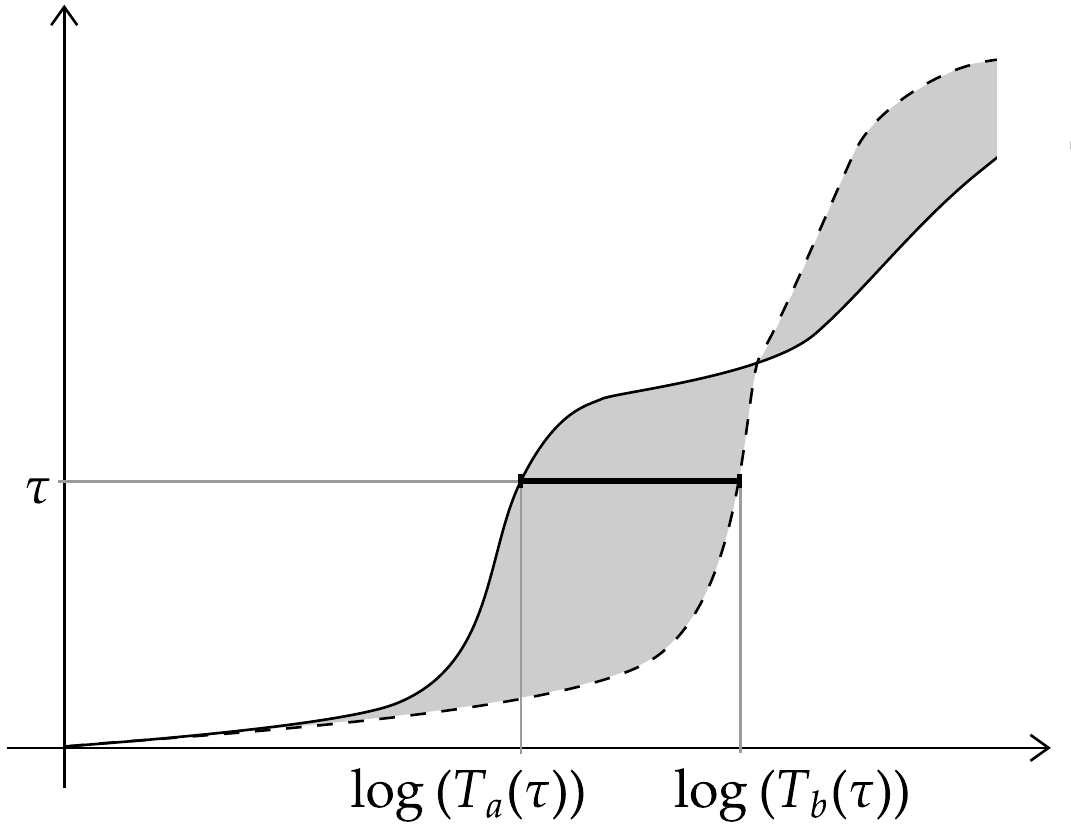}
    \caption{ERDs corresponding to algorithms $a$ (solid line) and $b$ (dashed line). The area between the lines (in gray) represents the difference in algorithm performance, $\Delta(a,b)$.}
    \label{fig:2}
\end{figure}

According to~\cite{Bartz-Beielstein2020}, a good test suite should include problems that ``enable the user to
tell the algorithms apart in the performance space''. To measure the ability of a given problem to differentiate among MOEAs, we rely on the area between the corresponding ERDs (see Fig.~\ref{fig:2}, area in gray). The intuition is that a large area between ERDs indicates large differences between the runtimes and in turn between the algorithms. 

Based on the area between the two ERDs we want to propose a metric, $\Delta$, that with a single number provides information about the similarity between algorithms and their performance. Let us assume we are dealing with two algorithms $a$ and $b$, and we have their corresponding runtime data sets for a certain optimization problem $\{T_a(\tau)\}_{\tau}$ and $\{T_b(\tau)\}_{\tau}$. Then, the area of a single segment between the runtimes (in the logarithmic scale) for the same target can be obviously calculated as follows (see Fig.~\ref{fig:2}, the bold line between runtimes):
\begin{equation} \label{eq:segment}  \frac{\abs{\log\left(T_a(\tau)\right) - \log\left(T_b(\tau)\right)}}{\ntarget{}},
\end{equation}
where $\ntarget{}$ is the number of targets. In particular, when a certain runtime is missing, we set its value to the maximal budget (number of function evaluations). This is done for calculation purpose and has no particular meaning. Using the formula (\ref{eq:segment}), the area bounded by the two ERDs and thus $\Delta$ can be expressed as the sum of these segment areas over all the targets:
\begin{equation}
    \Delta(a, b) = \frac{\sum_{\tau \in \tau}\abs{\log\left(\frac{T_a(\tau)}{T_b(\tau)}\right)}}{\ntarget{} \log N_{f}},
\end{equation}
where $N_{f}$ is the number of function evaluations. The formula is additionally divided by $\log N_f$ for normalization purposes. It can be easily seen that $\Delta(a, b) \in [0,1]$ for all algorithms and problems. In particular, small values indicate similar behavior of the chosen algorithms, and vice versa. For example, in the extreme case algorithm $a$ might solve all the targets within a single evaluation, while the algorithm $b$ does not reach any target. In this case $\Delta(a, b)=1$. In the second extreme case all the runtimes coincide and $\Delta(a, b) = 0$.

The $\Delta(a, b)$ metric can be additionally expressed as:
\begin{equation}
    \Delta(a, b) = \frac{\ntarget{-}}{\ntarget{}} \Delta^-(a, b) + \frac{\ntarget{+}}{\ntarget{}} \Delta^+(a, b)
\end{equation}
where $\Delta^-$ and $\Delta^+$ represent the sum of segments (\ref{eq:segment}) over targets measuring constraint satisfaction, $\tau^-$, and targets expressing the algorithm effectiveness in approximating the Pareto front (called front approximation for short), $\tau^+$, respectively. Additionally, $\ntarget{+}$ is the number of $\tau^+$ targets, and $\ntarget{-}$ the number of $\tau^-$ targets. In particular, $\Delta^-$ can be seen as a measure of algorithm differences in constraint satisfaction, while $\Delta^+$ is a metric measuring differences in front approximation.

\section{Experimental Analysis} 
\label{sec:experiments}

This section introduces the test suites of CMOPs used for the experiments, discusses the chosen MOEAs and their CHTs, and provides the parameter and implementation details.

\subsection{Test Suites} 
\label{sec:test_suites}

The most notable artificial test suites of CMOPs used to assess the performance of constrained multiobjective optimization algorithms are CTP~\cite{Deb2001}, CF~\cite{Zhang2008}, C-DTLZ~\cite{Jain2014}, NCTP~\cite{Li2016}, DC-DTLZ~\cite{Li2019}, LIR-CMOP~\cite{Fan2019a}, DAS-CMOP~\cite{Fan2019b}, and MW~\cite{Ma2019}. The basic characteristics of the test suites are summarized in Table~\ref{tab:suites}. 

\begin{table}
    \centering
    \caption{Characteristics of test suites: number of problems, dimension of the search space $D$, number of objectives $M$, and number of constraints $I$.}
    \label{tab:suites}
    \begin{tabular}{lllll}
        \hline
        Test suite & \#problems & $D$ & $M$ & $I$ \\
        \hline
        CTP~\cite{Deb2001} & \phantom{0}8 & * & 2 & 2, 3 \\
        CF~\cite{Zhang2008} & 10 & * & 2, 3 & 1, 2 \\
        C-DTLZ~\cite{Jain2014} & \phantom{0}6 & * & * & 1, * \\
        NCTP~\cite{Li2016} & 18 & * & 2 & 1, 2 \\
        DC-DTLZ~\cite{Li2019} & \phantom{0}6 & * & * & 1, * \\
        DAS-CMOP~\cite{Fan2019b} & \phantom{0}9 & * & 2, 3 & 7, 11 \\
        LIR-CMOP~\cite{Fan2019a} & 14 & * & 2, 3 & 2, 3 \\ 
        MW~\cite{Ma2019} & 14 & * & 2, * & 1--4 \\
        \hline
        \multicolumn{5}{l}{*Scalable parameter.}\\
    \end{tabular}
\end{table}

\subsection{Multiobjective Evolutionary Optimization Algorithms} 
\label{sec:moea}
 
Three well-known CMOEAs were used to investigate the proposed assessment methodology and to compare the test suites: NSGA-III~\cite{Deb14, Jain2014}, C-TAEA~\cite{Li2019} and MOEA/D-IEpsilon~\cite{Fan2019a} all equipped with their default CHTs.

NSGA-III is a well-known algorithm that uses the constrained domination principle (CDP)~\cite{Deb02} as a CHT. This principle is an extension of the dominance relation and is the most widely-used technique to solve CMOPs. It strictly favors feasible solutions over infeasible ones. While feasible solutions are compared based on Pareto dominance ($\preceq$), infeasible solutions are compared according to the overall constraint violation. The formal definition of CDP, as presented in~\cite{Zhu2020}, is the following:
\begin{equation}
    x \preceq_{\mathrm{CDP}} y \Leftrightarrow
    \begin{cases}
    x \preceq y & \text{if } v(x) = v(y) = 0 \\
    v(x) < v(y) & \text{otherwise} \\
    \end{cases}.
\end{equation}

Next, the CHT used in MOEA/D-IEpsilon is based on the $\varepsilon$-constraint relation:
\begin{equation}
    x \preceq_{\varepsilon} y \Leftrightarrow
    \begin{cases}
    g^{\mathrm{tc}}(x) < g^{\mathrm{tc}}(y) & \text{if } (v(x) \leq \varepsilon \text{ and } v(y) \leq \varepsilon) \\
    & \text{or } v(x) = v(y) \\
    v(x) < v(y) & \text{otherwise} \\
    \end{cases},
\end{equation}
where 
\begin{equation}
    g^{\mathrm{tc}}(x \mid \nu)=\max_{1 \leq m \leq M} \{\nu_m\abs{f_m(x) - z^*_m}\}
\end{equation}
is the Tchebycheff aggregation function.

The comparison level $\varepsilon_t$ is updated in each generation following the expression:
\begin{equation} \label{eq:eps_update}
    \varepsilon_{t} =
    \begin{cases}
    v(x^{\theta}) & \text{if } t = 0 \\
    (1 - \tau) \varepsilon_{t - 1} & \text{if } \rho_{\mathrm{F}}(P_t) < \alpha \text{ and } t < T_{\mathrm{c}} \\
    (1 + \tau) v_{\mathrm{max}} & \text{if } \rho_{\mathrm{F}}(P_t) \geq \alpha \text{ and } t < T_{\mathrm{c}} \\
    0 & \text{if } t \geq T_{\mathrm{c}} \\
    \end{cases}
\end{equation}
where $t$ is the generational counter, $\tau$, $\alpha$, $T_{\mathrm{c}}$ are user-defined parameters, $v(x^{\theta})$ is the overall constraint violation of the top $\theta$-th individual (according to overall constraint violation value) in the initial population, and $\rho_{\mathrm{F}}(P_t)$ is the proportion of feasible solutions in the current population $P_t$. 
Additional details on MOEA/D-IEpsilon can be found in~\cite{Fan2019a}.

Finally, the main idea behind C-TAEA is the maintenance of two separate archives. One archive is used to promote convergence, while the other one to maintain diversity. Besides, a special restricted mating approach is employed to balance between the two archives. The CHT used by C-TAEA is incorporated within the update of the convergence archive. Similarly to CDP, this CHT strictly favors feasible solutions which are compared based on Pareto dominance. On the other hand, the infeasible solutions are ranked using nondominated sorting for a custom bi-objective problem expressed as
\begin{equation} \label{eq:cht_ctaea}
    \text{minimize} \quad (v(x), g^{\mathrm{tc}}(x \mid \nu)).
\end{equation}
The convergence archive is updated with all feasible solutions and the best infeasible solutions according to the Pareto ranking applied in~(\ref{eq:cht_ctaea}). In addition, the diversity archive does not consider feasibility at all allowing infeasible solutions to persists in the population. More information on this method are available in~\cite{Li2019}.

We chose three MOEAs with complementary CHTs: (i) the CDP employed in NSGA-III strictly favors feasible solutions, (ii) the diversity archive in C-TAEA allows infeasible solutions to remain in the population, and (iii) MOEA/D-IEpsilon adaptively updates the comparison level each generation following~(\ref{eq:eps_update}); when the feasibility ratio of the current population becomes large, $\varepsilon_t$ increases and progressively more solutions (infeasible ones included) are compared solely according to the objective values.

\subsection{Parameter Settings} 
\label{sec:experimental_setup}

The proposed performance assessment is demonstrated on on the test suites listed in Table~\ref{tab:suites}. In particular, three-objective C-DTLZ and DC-DTLZ problems were considered with the default number of constraints. Additionally, a difficulty triplet of (0.5, 0.5, 0.5) was used for the DAS-CMOP suite as this is by far the most frequently used difficulty triplet in the literature.

Three dimensions of the search space $D \in \{5, 10, 30\}$ were used to evaluate the proposed performance assessment methodology and compare the test suites. All the Pareto fronts can be analytically expressed and the corresponding hypervolume values exactly calculated.

\begin{table}
    \centering
    \caption{The population size $N_{\mathrm{p}}$ and number of generations $N_{\mathrm{g}}$ used in the experimental analysis, based on the number of objectives $M$ and the dimension of the search space $D$.}
    \label{tab:settings}
    \scriptsize
    \begin{tabular}{l@{\hspace*{0.035\textwidth}}lll@{\hspace*{0.035\textwidth}}lll}
        \hline
        & \multicolumn{3}{c}{$M=2$} & \multicolumn{3}{c}{$M=3$} \\ 
        \cline{2-7}
        $D$ & \phantom{00}5 & \phantom{0}10 & \phantom{00}30 & \phantom{00}5 & \phantom{0}10 & \phantom{00}30 \\
        \hline
        $N_{\mathrm{p}}$ & 200 & 200 & \phantom{0}200 & 300 & 300 & \phantom{0}300 \\
        $N_{\mathrm{g}}$ & 300 & 600 & 1800 & 200 & 400 & 1200 \\
        \hline
    \end{tabular}
\end{table}

All MOEAs were run with an equal population size, $N_{\mathrm{p}}$, and the same number of generations, $N_{\mathrm{g}}$. In particular, the population size was set to $N_{\mathrm{p}} = 100M$. The number of generations was set to $N_{\mathrm{g}} = 120D/M$ and was selected as approximately the minimal value to obtain convergence for all the MOEAs (in total, $12000D$ function evaluations). Note, the division by $M$ in the expression for $N_{\mathrm{g}}$ is necessary to enable aggregation over CMOPs with different number of objectives. The resulting values for $N_{\mathrm{p}}$ and $N_{\mathrm{g}}$ are shown in Table~\ref{tab:settings}. 

Other parameters of the algorithms and their operators were set to their default values~\cite{Deb14, Li2019, Fan2019a}: The polynomial mutation was used in all the MOEAs. The mutation probability was set to $1/D$ and the distribution index to 20. Specifically, the simulated binary crossover was used in NSGA-III and C-TAEA with the crossover probability of 1 and the distribution index of 30. In contrast, a differential-evolution-based crossover was used in MOEA/D with a crossover probability of 1 and scaling factor of 0.5. Additionally, in MOEA/D, the neighborhood size was set to 30, probability of neighborhood mating to 0.9, the maximal number of solutions replaced by a child to 2, $\tau$ to 0.1, $\alpha$ to 0.95, $T_{\mathrm{c}}$ to 0.8, and $\theta$ to $0.05N_{\mathrm{p}}$.

Finally, ERDs were computed without employing restarts or bootstrapping. 

\subsection{Target Precision Values for CMOPs}
The values of the distance metric $d$ defined in (\ref{eq:dist}), as well as those of the overall constraint violation $v$ can result in different magnitudes for different CMOPs. Consequently, it is impossible to define a set of target precision values that would provide meaningful results for all the studied CMOPs. As we wanted to compare different CMOPs and suites, we first sampled 100 solutions $\{x^i\}_i$ for each CMOP and normalized $d$ and $v$ as follows:
\begin{equation}
    \widetilde{d} = d /10^{\lceil\log(\text{med}(\{d^i\}_i))\rceil}
\end{equation}
and
\begin{equation}
    \widetilde{v} = d /10^{\lceil\log(\text{med}(\{v^i\}_i))\rceil}
\end{equation}
where $\text{med}(\{d^i\}_i)$ and $\text{med}(\{v^i\}_i)$ are median values of the sets $\{d^i\}_i=\{d(x^i, Z)\}_i$ and $\{v^i\}_i=\{v(x^i)\}_i$, respectively.

After applying this procedure and preforming several experiments, we set $\tau^*$ to 1. Additionally, a good set of target precision values for $I^\mathrm{CMOP}$ corresponds to $\tau(\varepsilon) = \tau^\mathrm{ref} + \varepsilon$, where $\varepsilon \in \{10^k \mid k \in \{-5, -4.9, \dots, 0\}\} \cup \{1 + 10^k \mid k \in \{-5, -4.9, \dots, 0\}\}$. The first half of target precision values $\tau^+$ applies to feasible solutions and represents how well the algorithm approximates the Pareto front, while the second half of the targets $\tau^-$ is used to understand the algorithm performance in satisfying the constraints.

\subsection{Implementation Details}
\label{sec:implementation_details}

All the CMOPs, MOEAs and the performance measurement procedure were implemented in the Python programming language~\cite{Rossum09}. We used the \texttt{pymoo}~\cite{Blank20} implementation for CTP, DAS-CMOP, MW, NSGA-III, C-TAEA and hypervolume calculation. The rest of the suites, MOEA/D-IEpsilon and other functionalities were implemented from scratch.

\section{Results and Discussion}
\label{sec:results}

In this section, we first present the experimental results. Next, we discuss the existing test suites of CMOPs and their potential in differentiating algorithms. Finally, we present some limitations of the proposed methodology.

\subsection{Results}

\begin{figure*}[!t]
    \centering
    \subfloat[MW12 ($D=5$)]{\includegraphics[clip, trim={0, 10pt, 0, 25pt}, width=0.33\textwidth]{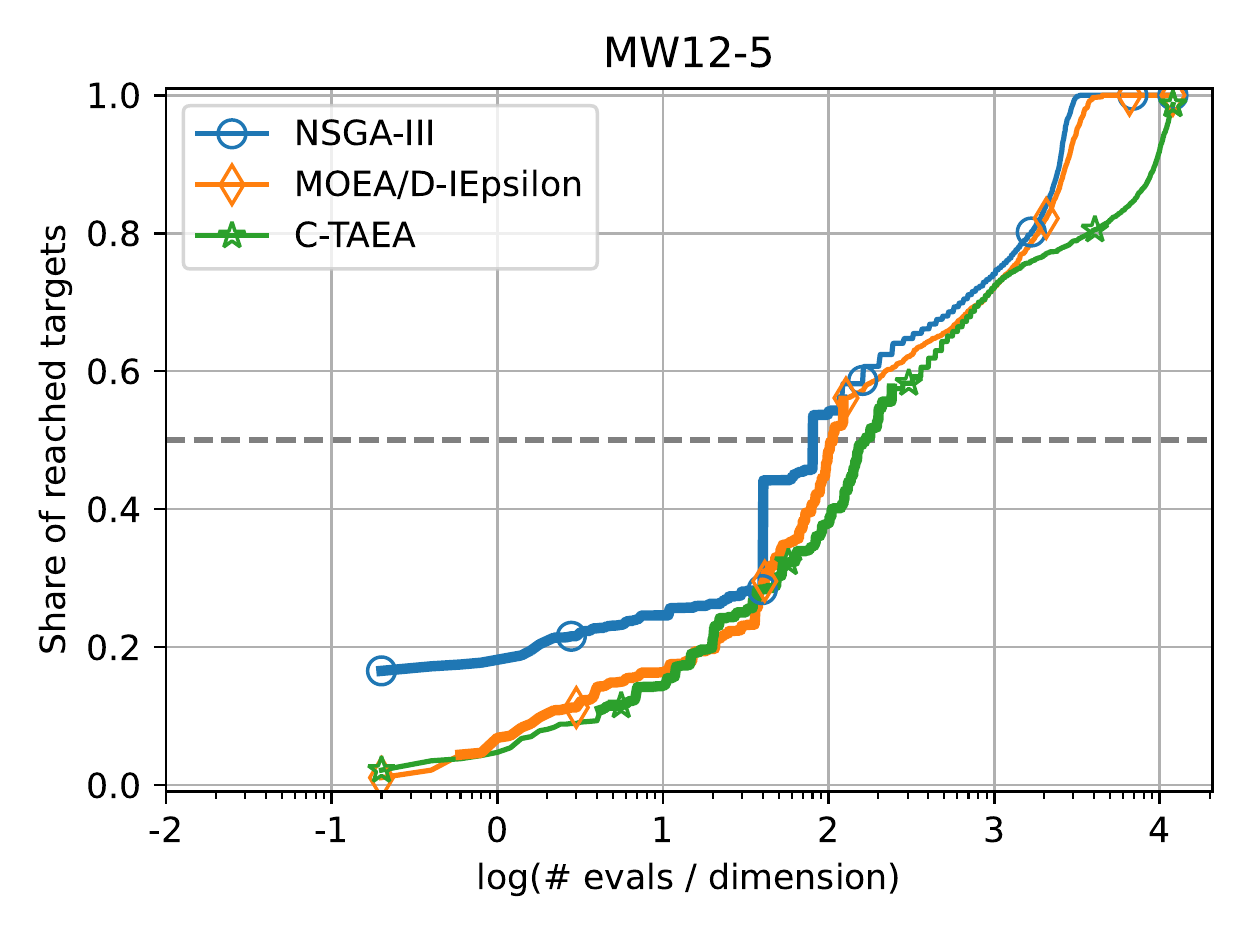}%
    \label{fig:ecdfs_mw12_d5}}
    \hfil
    \subfloat[MW12 ($D=10$)]{\includegraphics[clip, trim={0, 10pt, 0, 25pt}, width=0.33\textwidth]{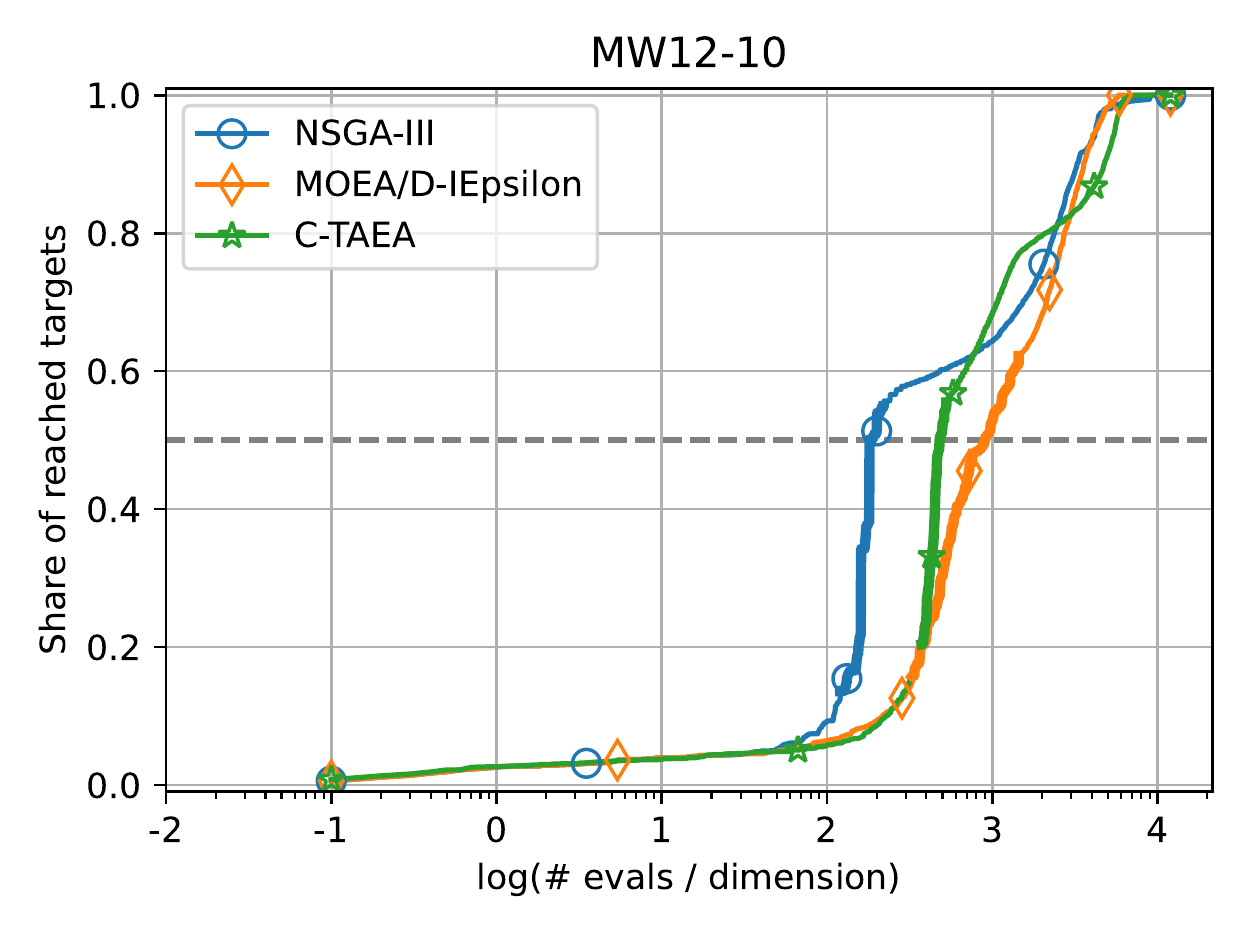}%
    \label{fig:ecdfs_mw12_d10}}
    \hfil
    \subfloat[MW12 ($D=30$)]{\includegraphics[clip, trim={0, 10pt, 0, 25pt}, width=0.33\textwidth]{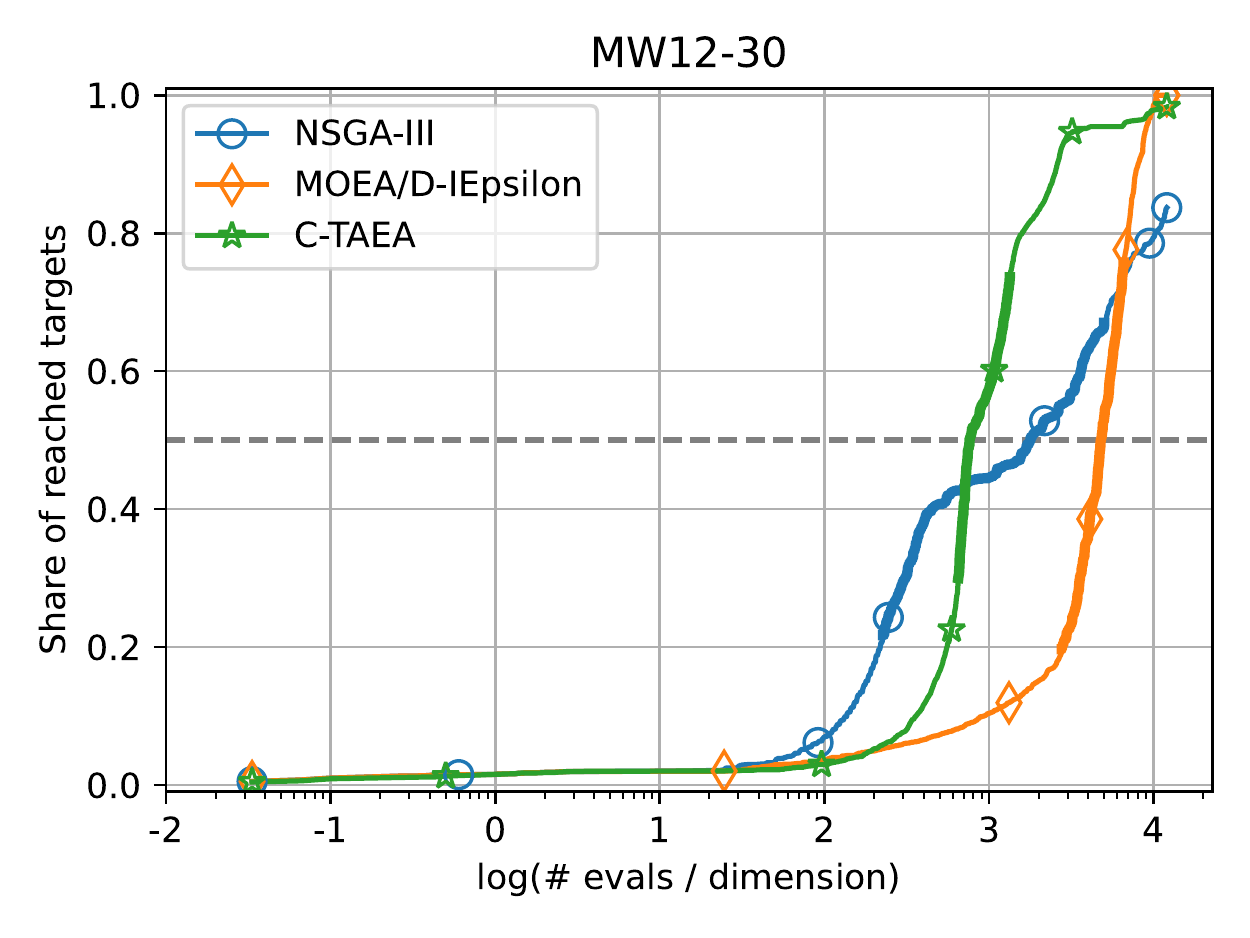}%
    \label{fig:ecdfs_mw12_d30}}
    \caption{Empirical runtime distribution aggregated over multiple runs for the MW12 problem for the three MOEAs and dimension 5 (left), 10 (center) and 30 (right).}
    \label{fig:ecdfs_mw13}
\end{figure*}

Let us first look at the results for a single problem. As an example we select the MW13 problem. Fig.~\ref{fig:ecdfs_mw13} shows the ERDs for this problem aggregated over 30 runs for each of the three algorithms. To aid visualization and comparison, the function evaluations ($x$-axis) are divided by problem dimension and shown in logarithmic scale. 

The horizontal dashed line divides the targets into $\tau^-$ and $\tau^+$. Note that this intuition is true only for a single run without aggregation. Nevertheless, if an ERD (single or aggregated) starts above this line, then all the algorithm runs started in the feasible region---the first initialized solution is feasible. On the other hand, if an ERD starts below the dashed line and never crosses it, then this indicates that the corresponding algorithm runs never reached the feasible region. Moreover, the line is thicker when some, but not all runs have found feasible solutions.

As we can see, all the MOEAs reach all the targets for the two smaller dimensions, while NSGA-III fails to reach some targets on 30-$D$ problems. All the algorithms are able to find feasible solutions in all of the runs with NSGA-III being generally quickest in this regard. 

\begin{figure*}[!t]
    \centering

    \subfloat[CTP ($D=5$)]{\includegraphics[width=0.33\textwidth]{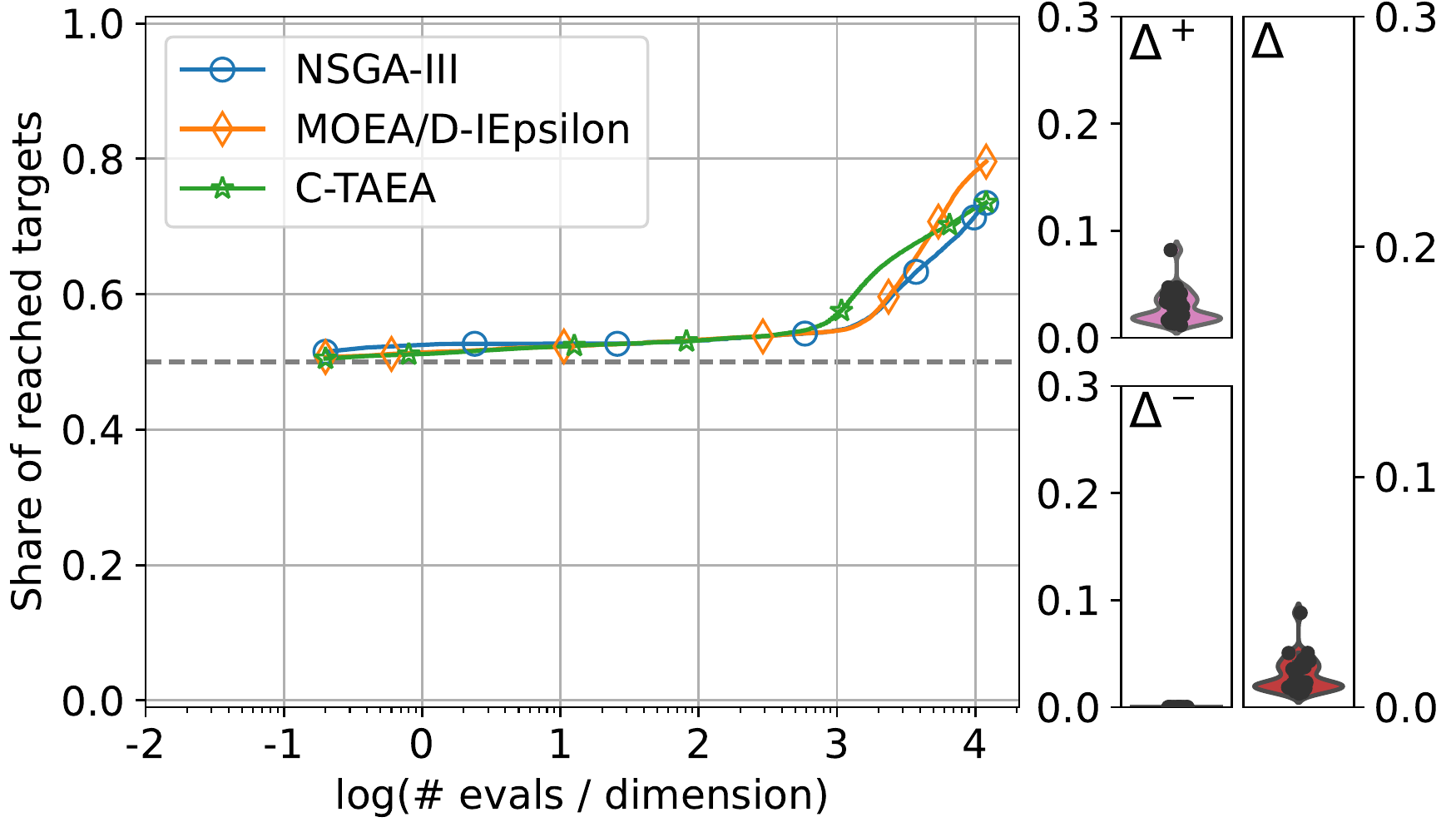}%
    \label{fig:CTP_d5}}
    \hfil
    \subfloat[CTP ($D=10$)]{\includegraphics[width=0.33\textwidth]{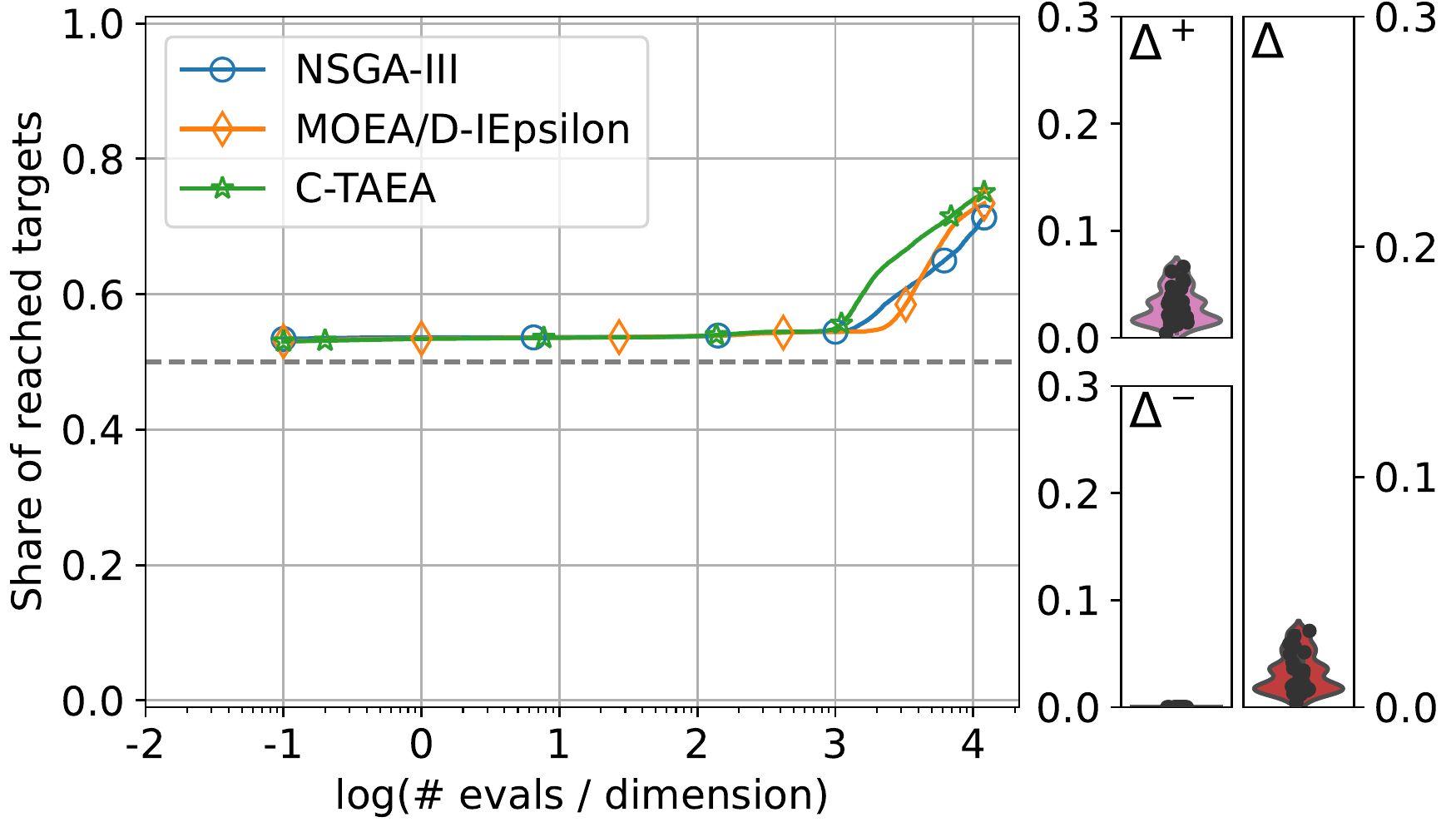}%
    \label{fig:CTP_d10}}
    \hfil
    \subfloat[CTP ($D=30$)]{\includegraphics[width=0.33\textwidth]{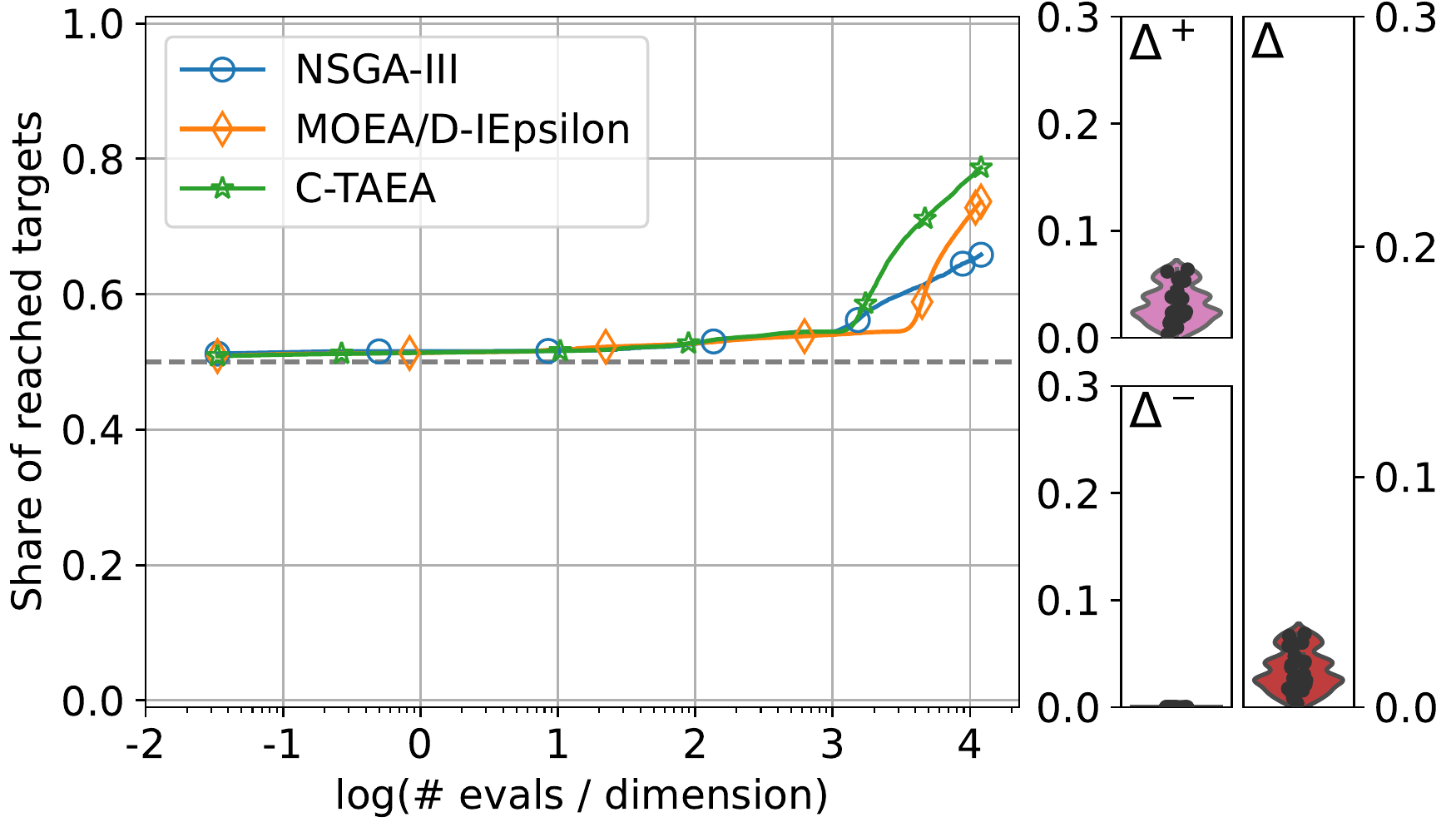}%
    \label{fig:CTP_d30}}
    \hfil
    
    \subfloat[CF ($D=5$)]{\includegraphics[width=0.33\textwidth]{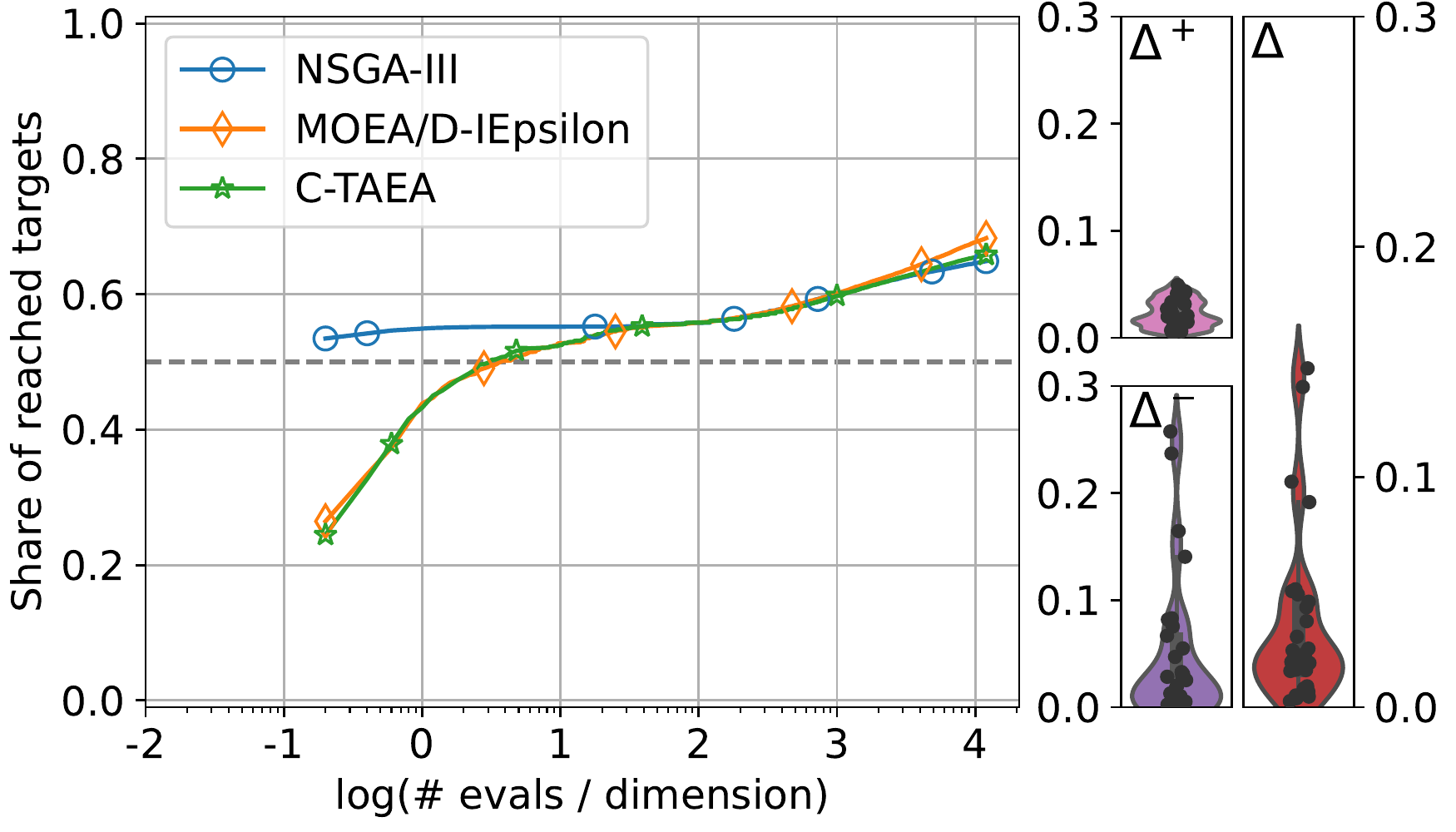}%
    \label{fig:CF_d5}}
    \hfil
    \subfloat[CF ($D=10$)]{\includegraphics[width=0.33\textwidth]{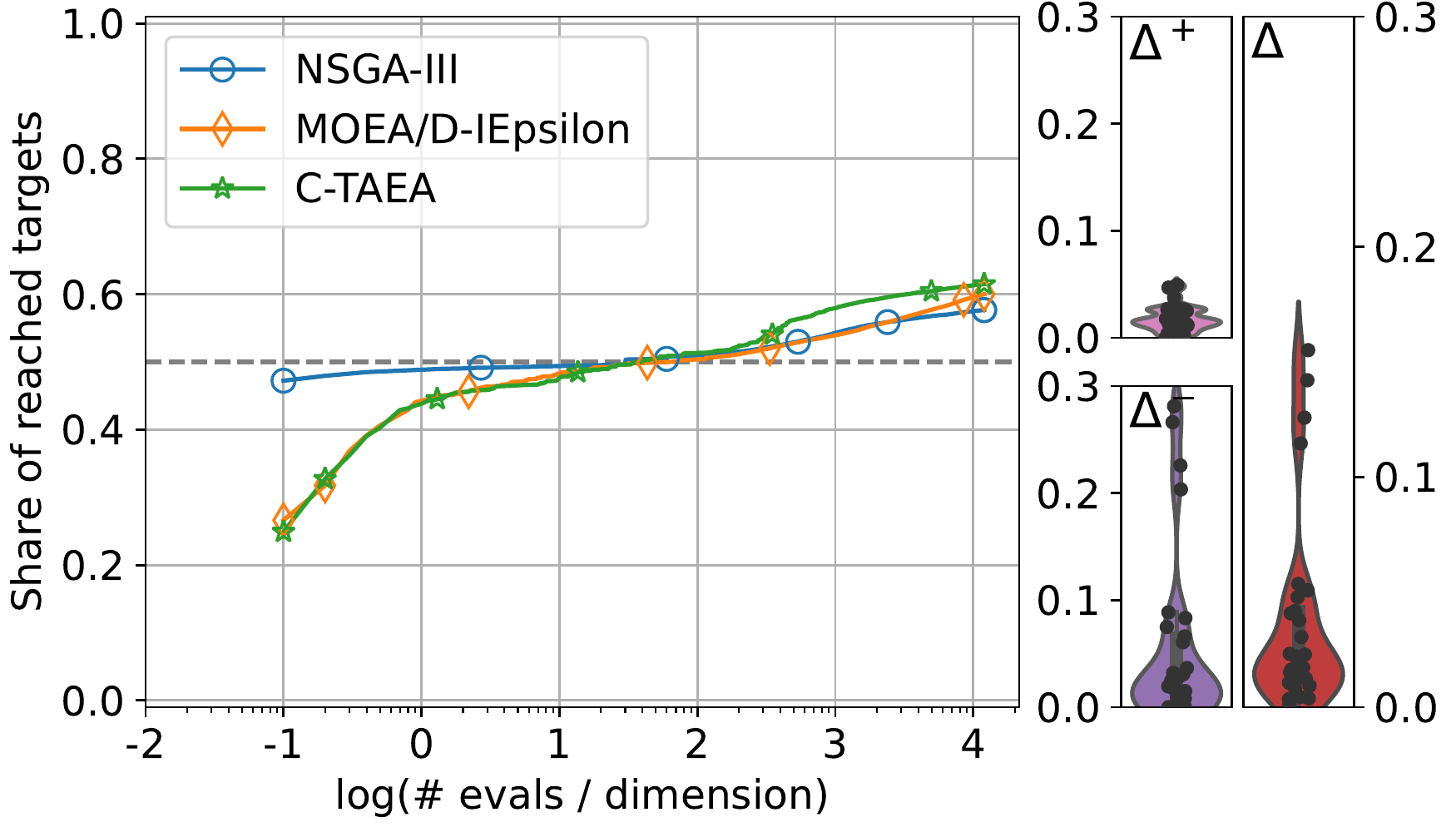}%
    \label{fig:CF_d10}}
    \hfil
    \subfloat[CF ($D=30$)]{\includegraphics[width=0.33\textwidth]{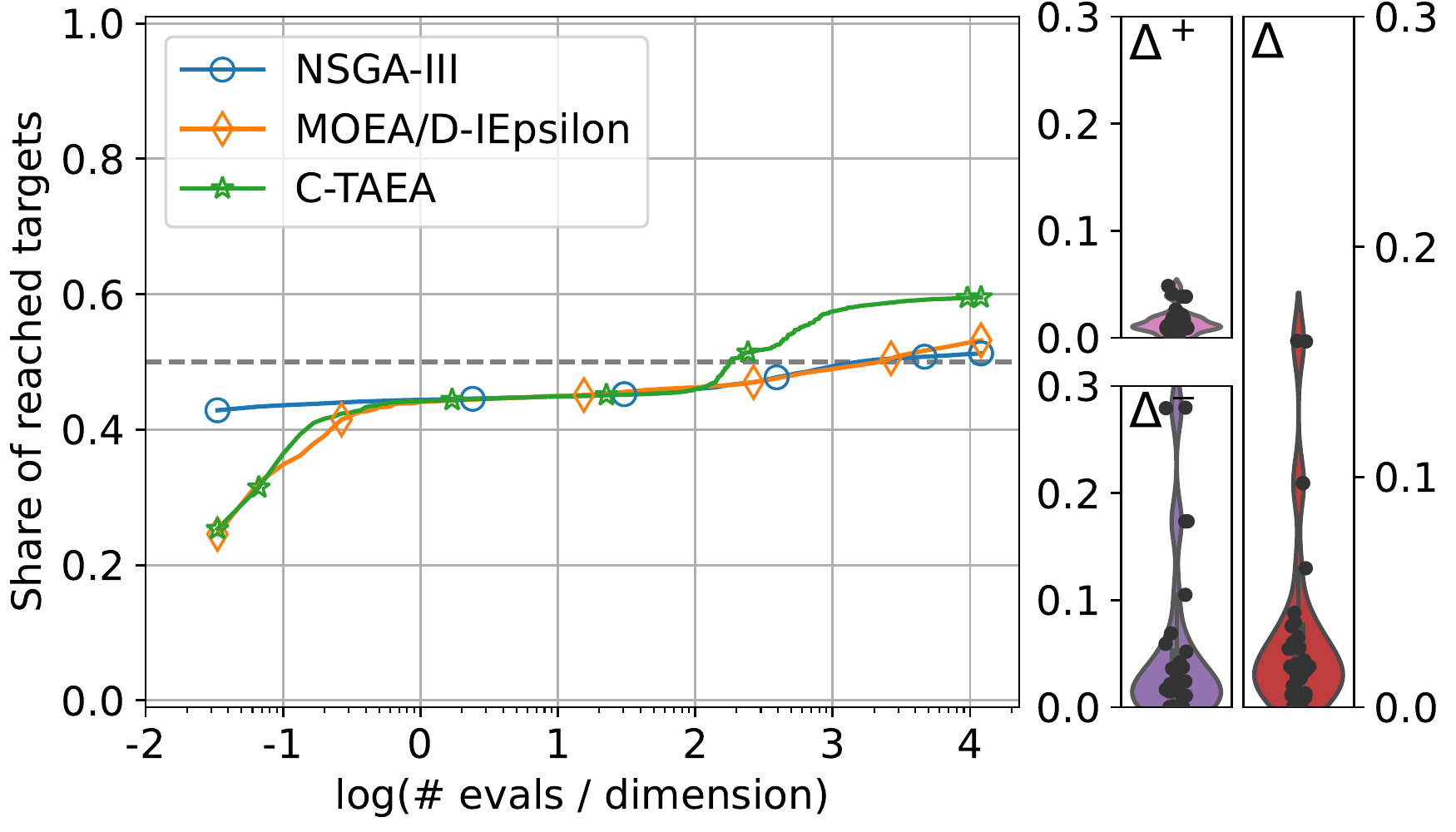}%
    \label{fig:CF_d30}}
    \hfil

    \subfloat[C-DTLZ ($D=5$)]{\includegraphics[width=0.33\textwidth]{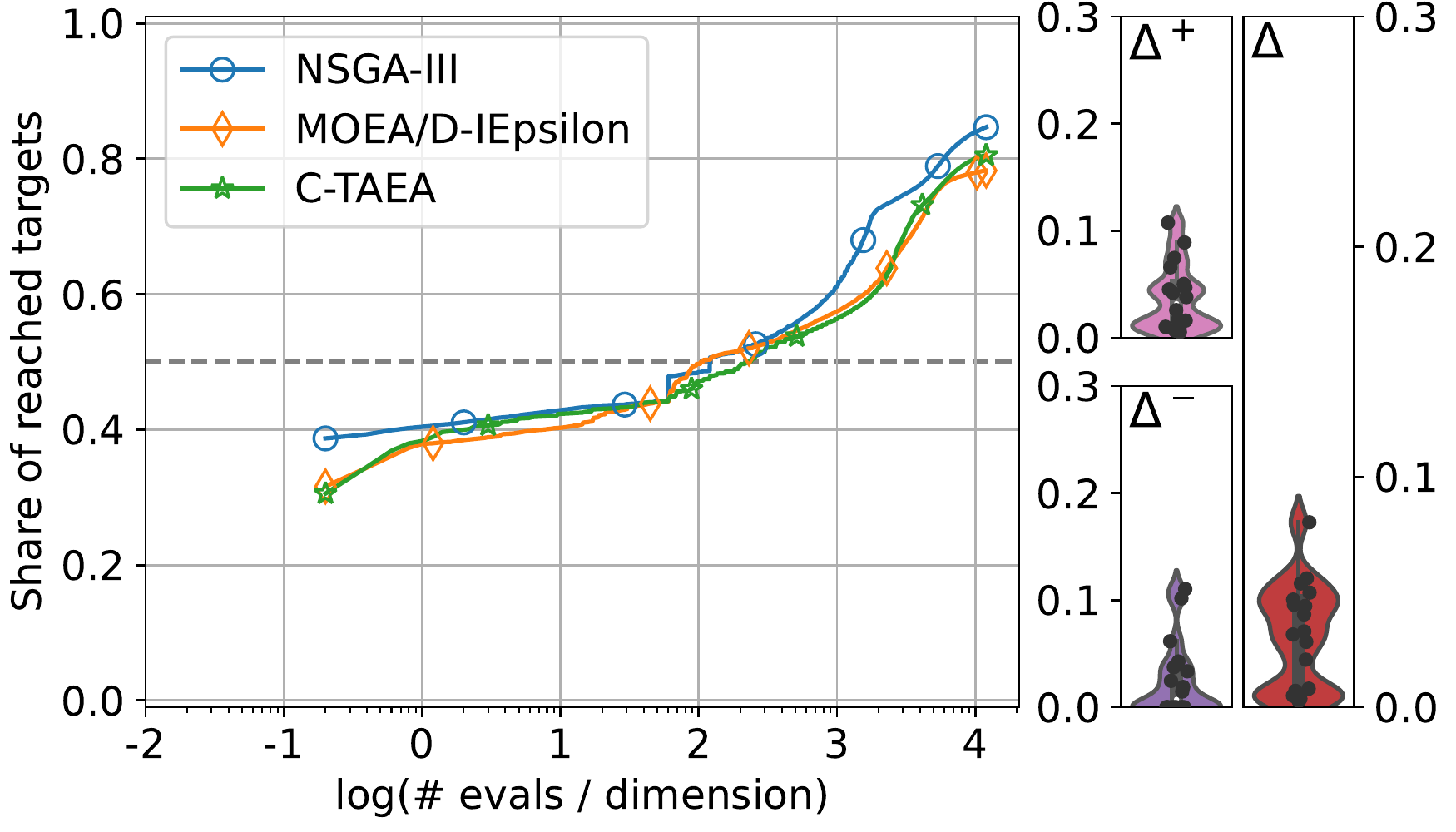}%
    \label{fig:C-DTLZ_d5}}
    \hfil
    \subfloat[C-DTLZ ($D=10$)]{\includegraphics[width=0.33\textwidth]{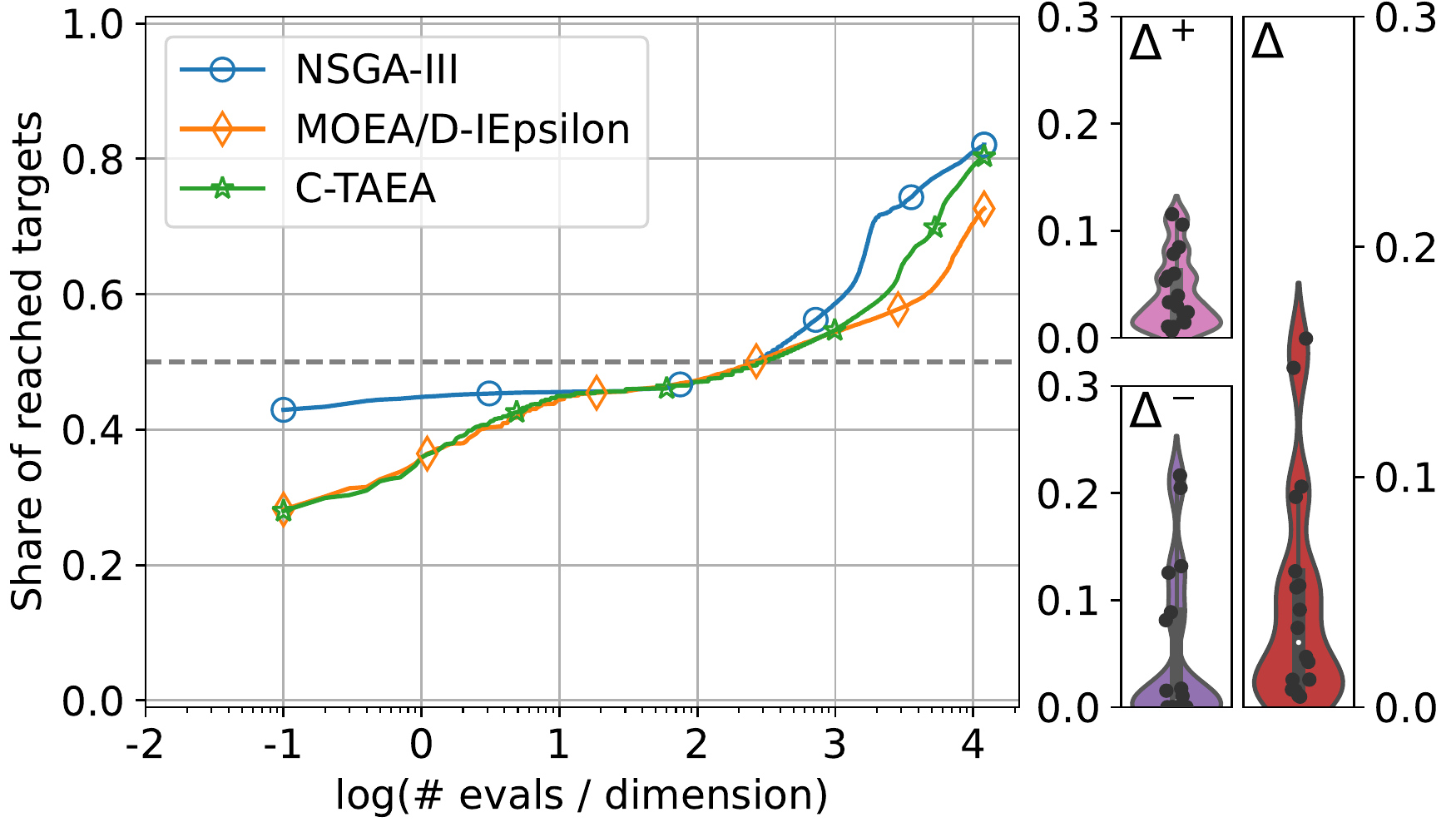}%
    \label{fig:C-DTLZ_d10}}
    \hfil
    \subfloat[C-DTLZ ($D=30$)]{\includegraphics[width=0.33\textwidth]{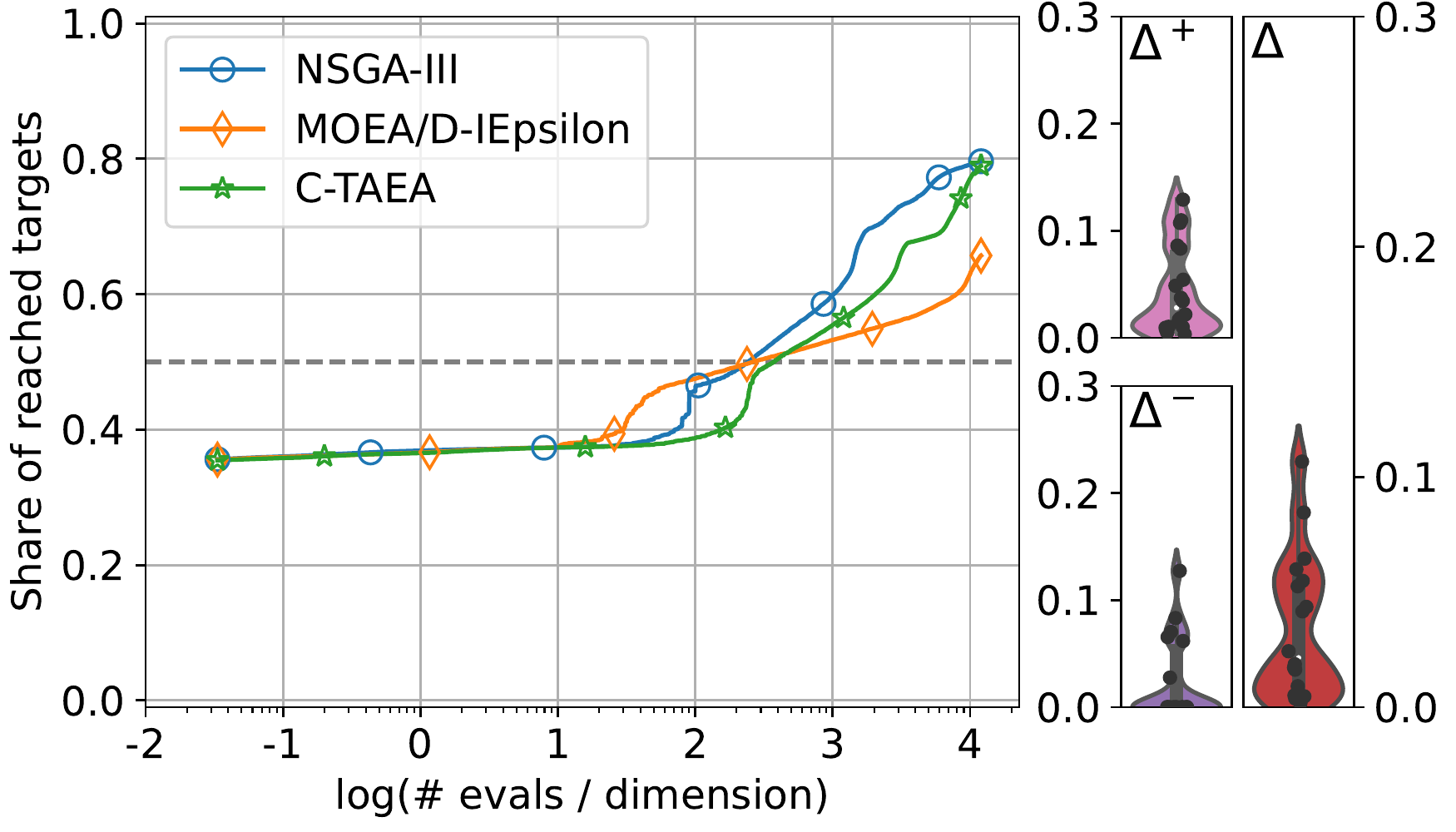}%
    \label{fig:C-DTLZ_d30}}
    \hfil

    \subfloat[NCTP ($D=5$)]{\includegraphics[width=0.33\textwidth]{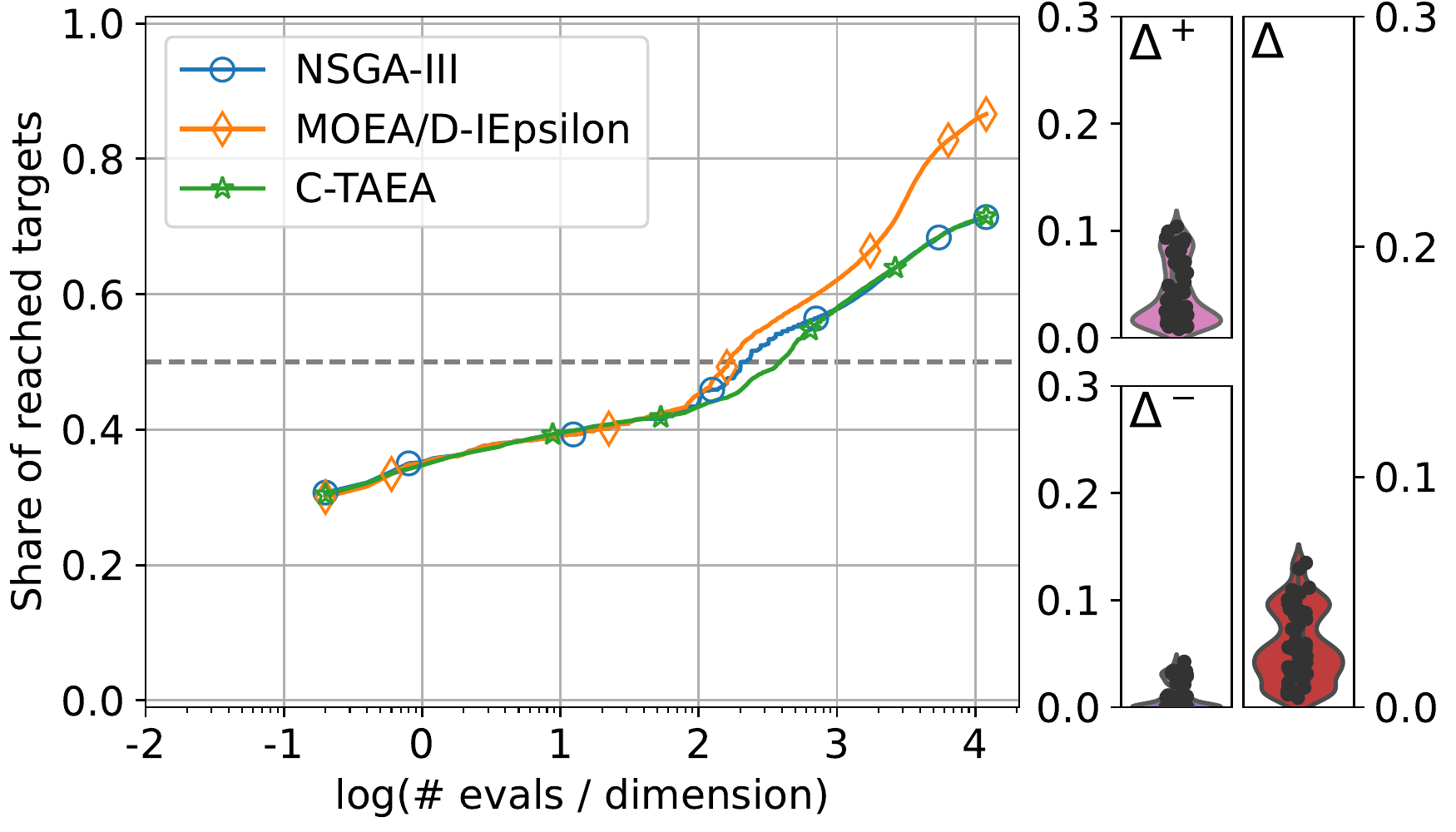}%
    \label{fig:NCTP_d5}}
    \hfil
    \subfloat[NCTP ($D=10$)]{\includegraphics[width=0.33\textwidth]{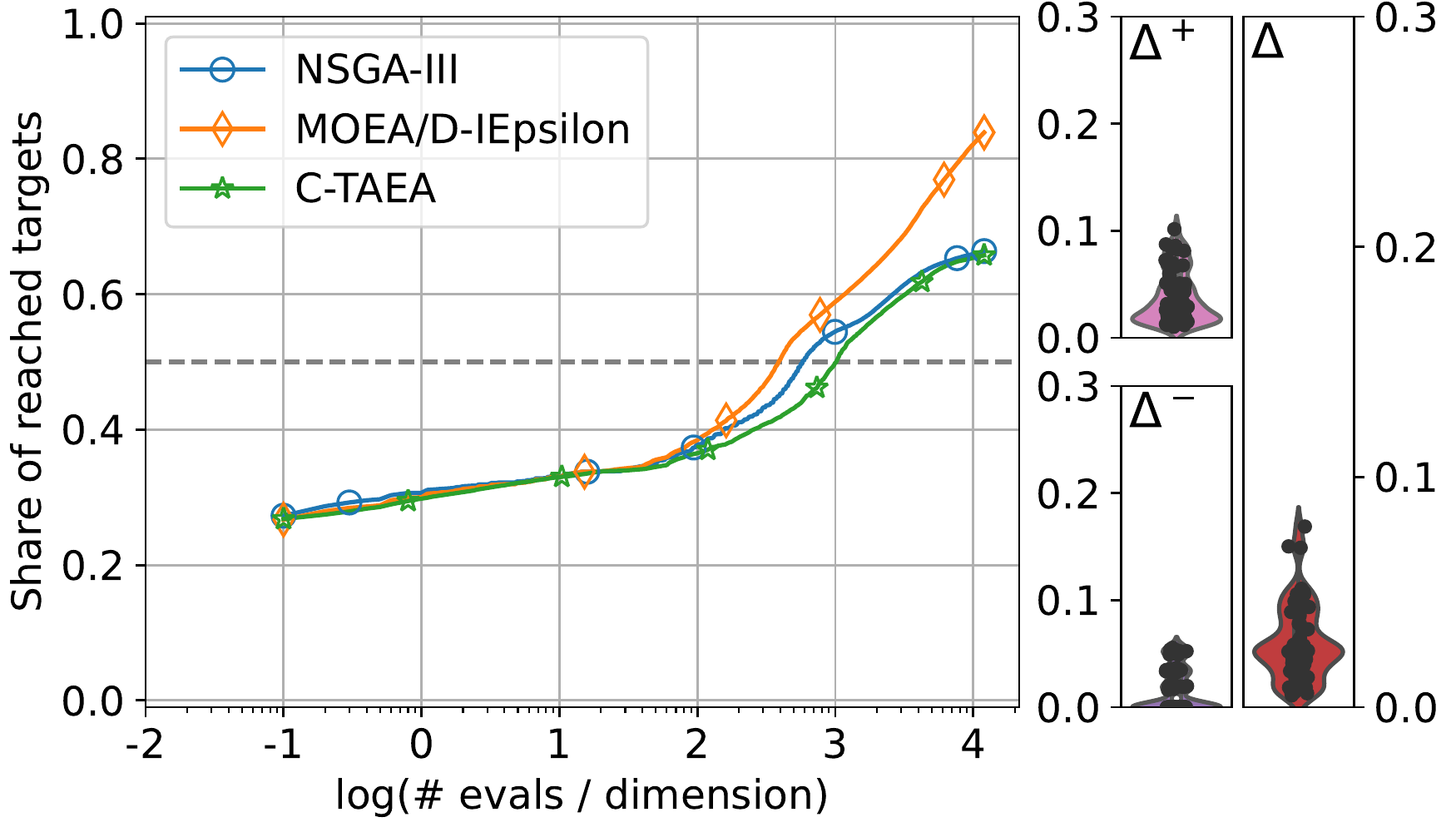}%
    \label{fig:NCTP_d10}}
    \hfil
    \subfloat[NCTP ($D=30$)]{\includegraphics[width=0.33\textwidth]{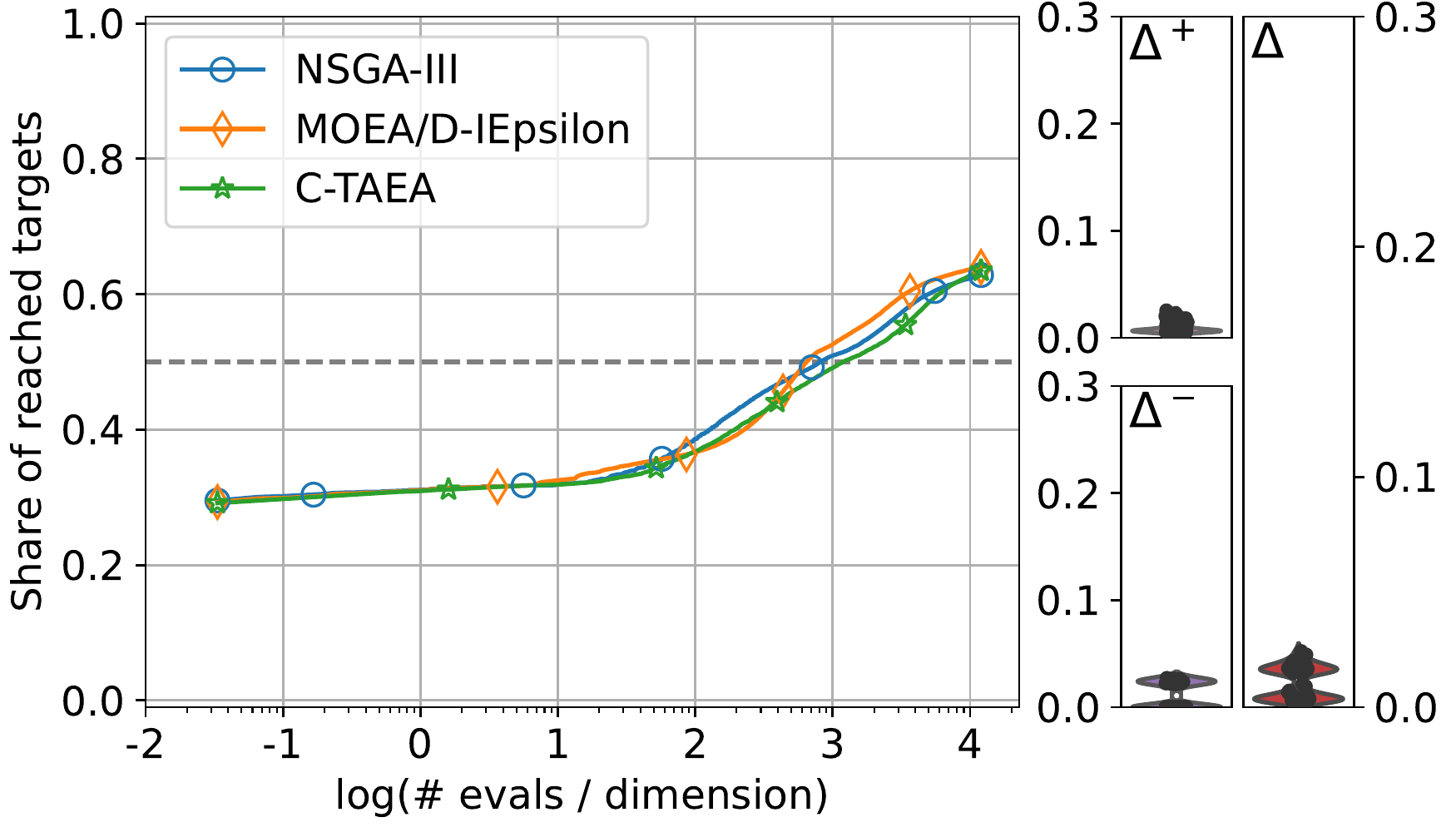}%
    \label{fig:NCTP_d30}}
    \hfil

    \caption{Results of the three MOEAs on CMOPs from CTP, CF, C-DTLZ, and NCTP suites. 
    The left plot of each subfigure shows empirical runtime distribution aggregated over all CMOPs in the suite and all targets in dimension 5 (left), 10 (center) and 30 (right). On the right of each subfigure, violin plots depict distributions of $\Delta^+$ (top left), $\Delta^-$ (bottom left), and $\Delta$ (right) values. The larger the diversity, the better.}
    \label{fig:ecdfs_1}
\end{figure*}

\begin{figure*}[!t]
    \centering

    \subfloat[DC-DTLZ ($D=5$)]{\includegraphics[width=0.33\textwidth]{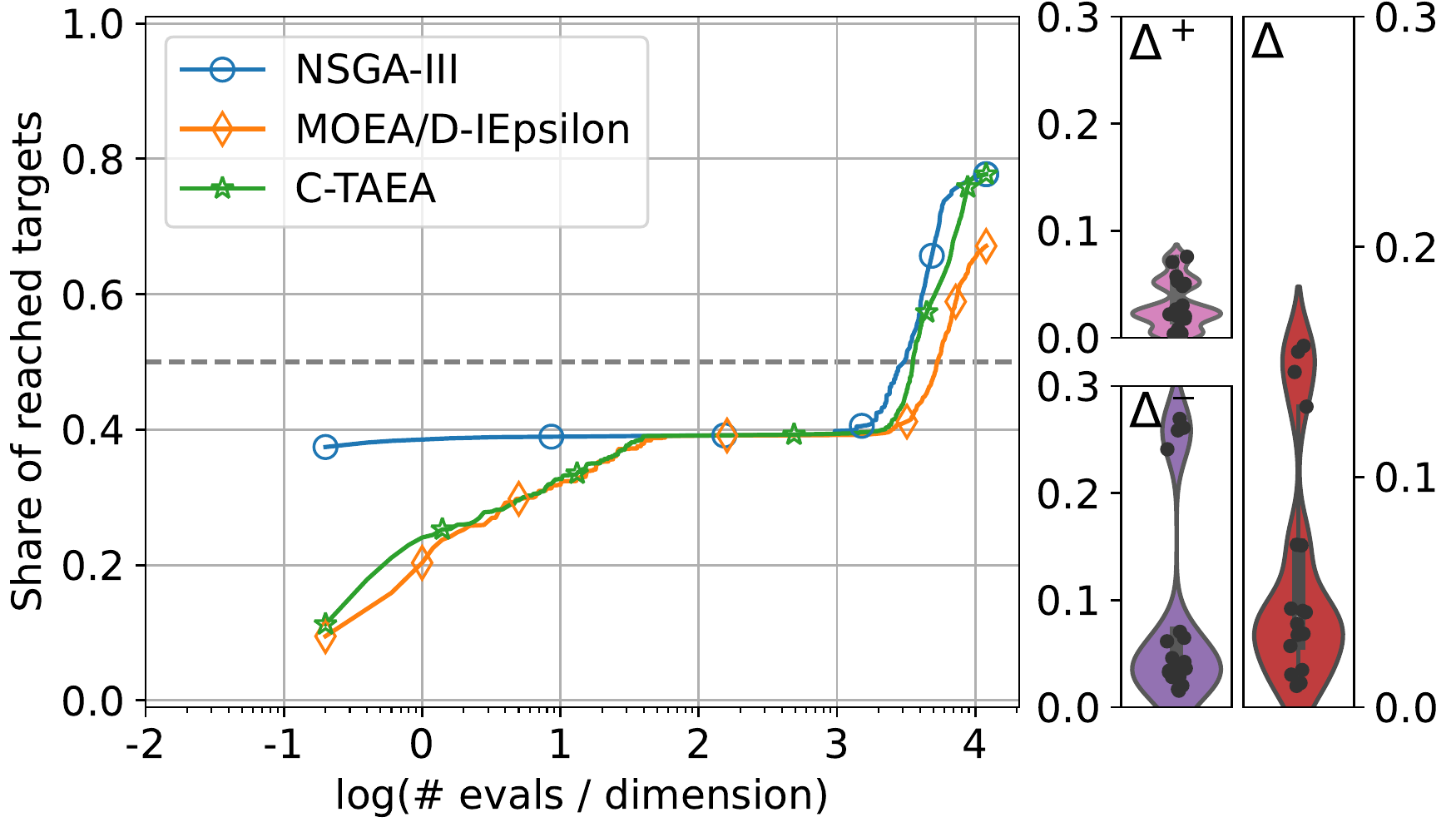}%
    \label{fig:DC-DTLZ_d5}}
    \hfil
    \subfloat[DC-DTLZ ($D=10$)]{\includegraphics[width=0.33\textwidth]{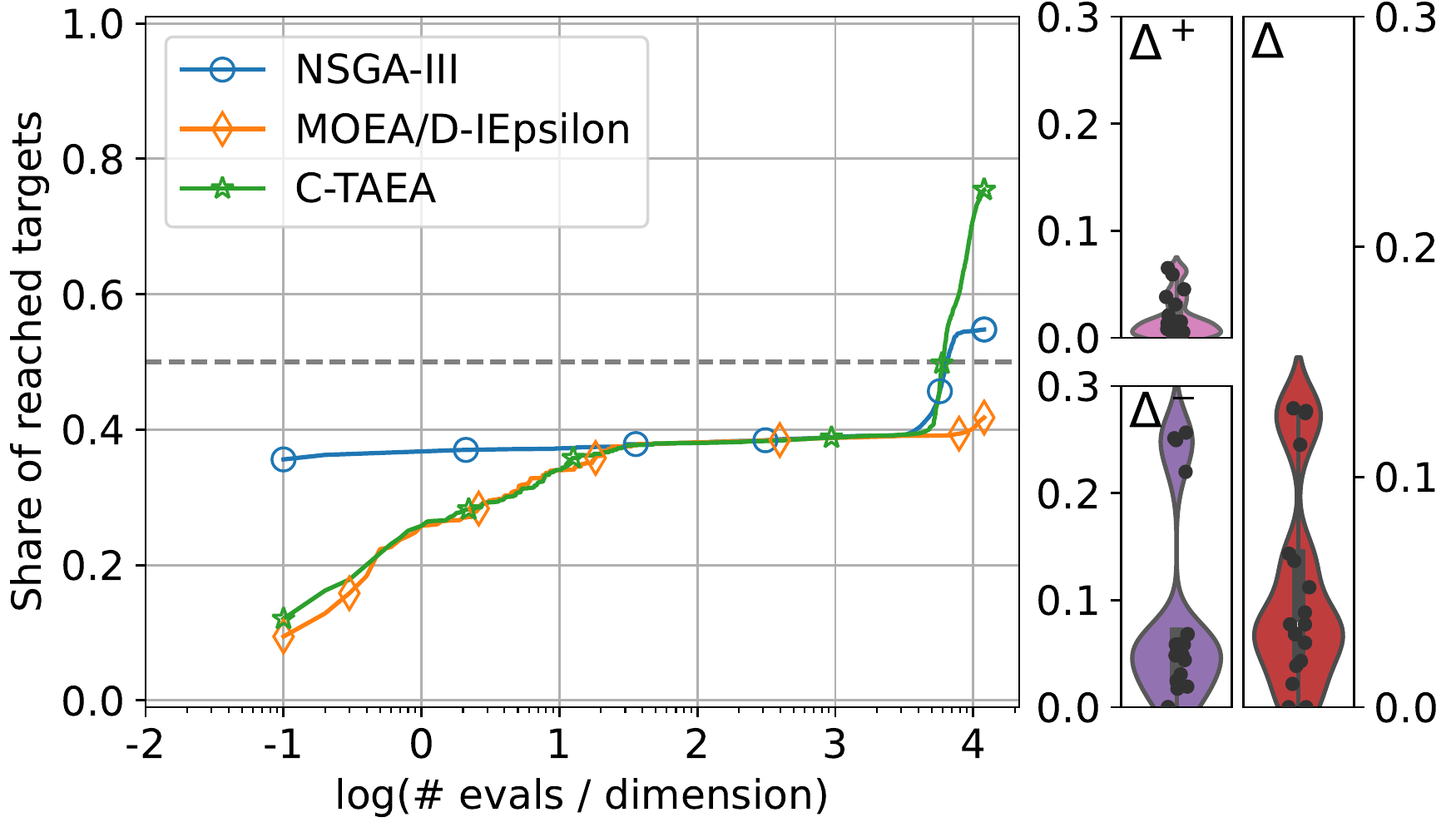}%
    \label{fig:DC-DTLZ_d10}}
    \hfil
    \subfloat[DC-DTLZ ($D=30$)]{\includegraphics[width=0.33\textwidth]{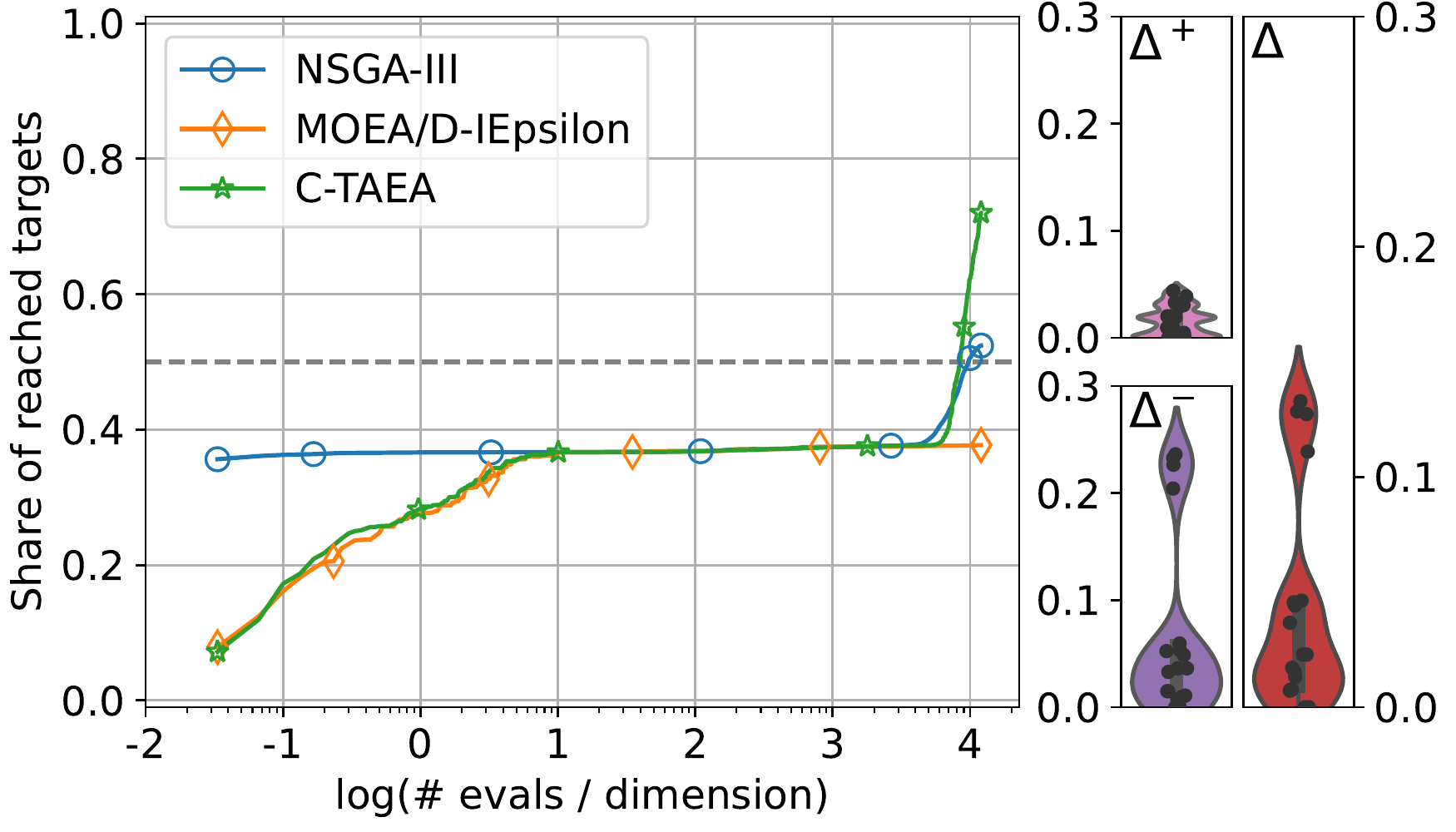}%
    \label{fig:DC-DTLZ_d30}}
    \hfil

    \subfloat[DAS-CMOP ($D=5$)]{\includegraphics[width=0.33\textwidth]{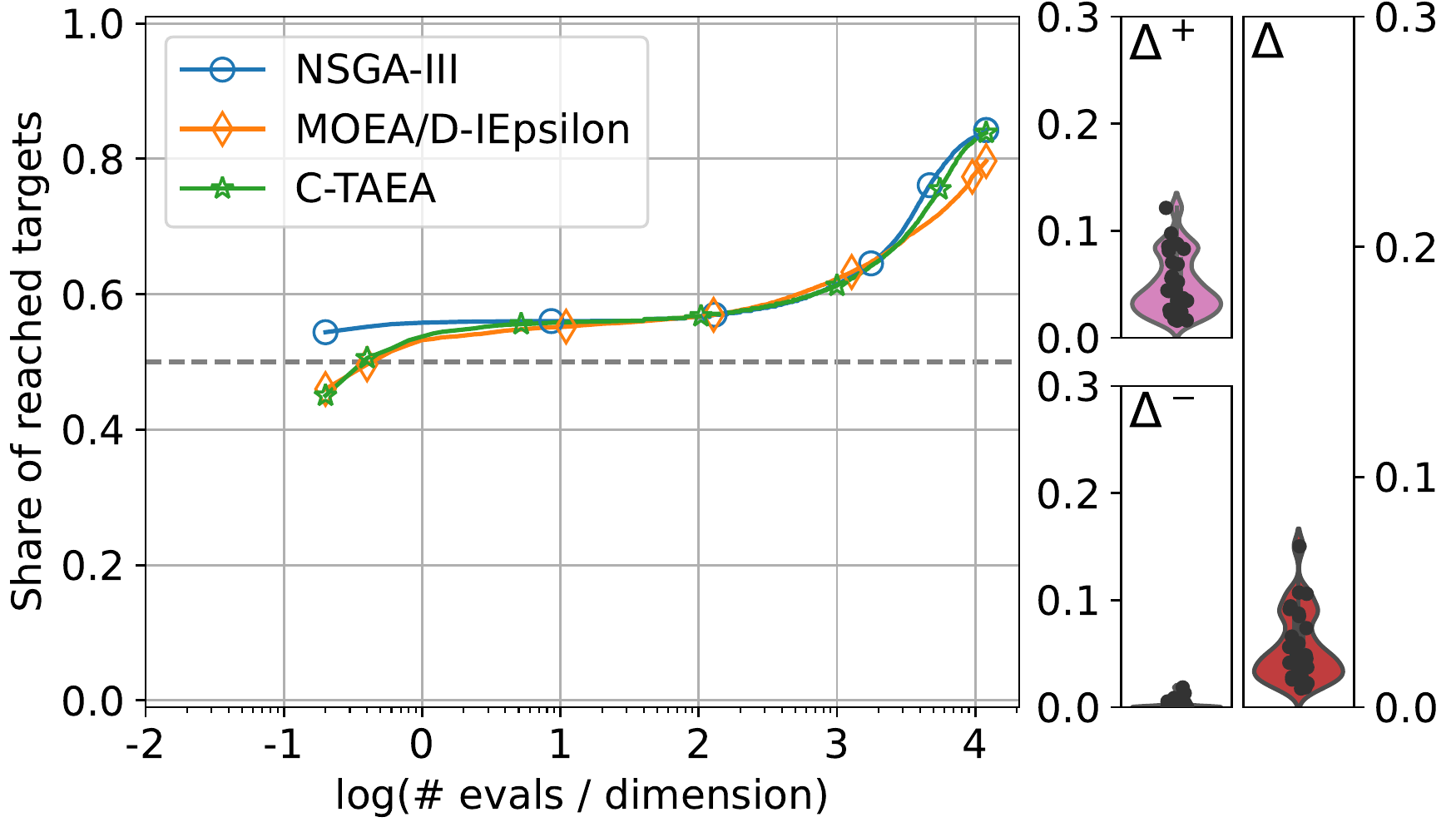}%
    \label{fig:DAS-CMOP_d5}}
    \hfil
    \subfloat[DAS-CMOP ($D=10$)]{\includegraphics[width=0.33\textwidth]{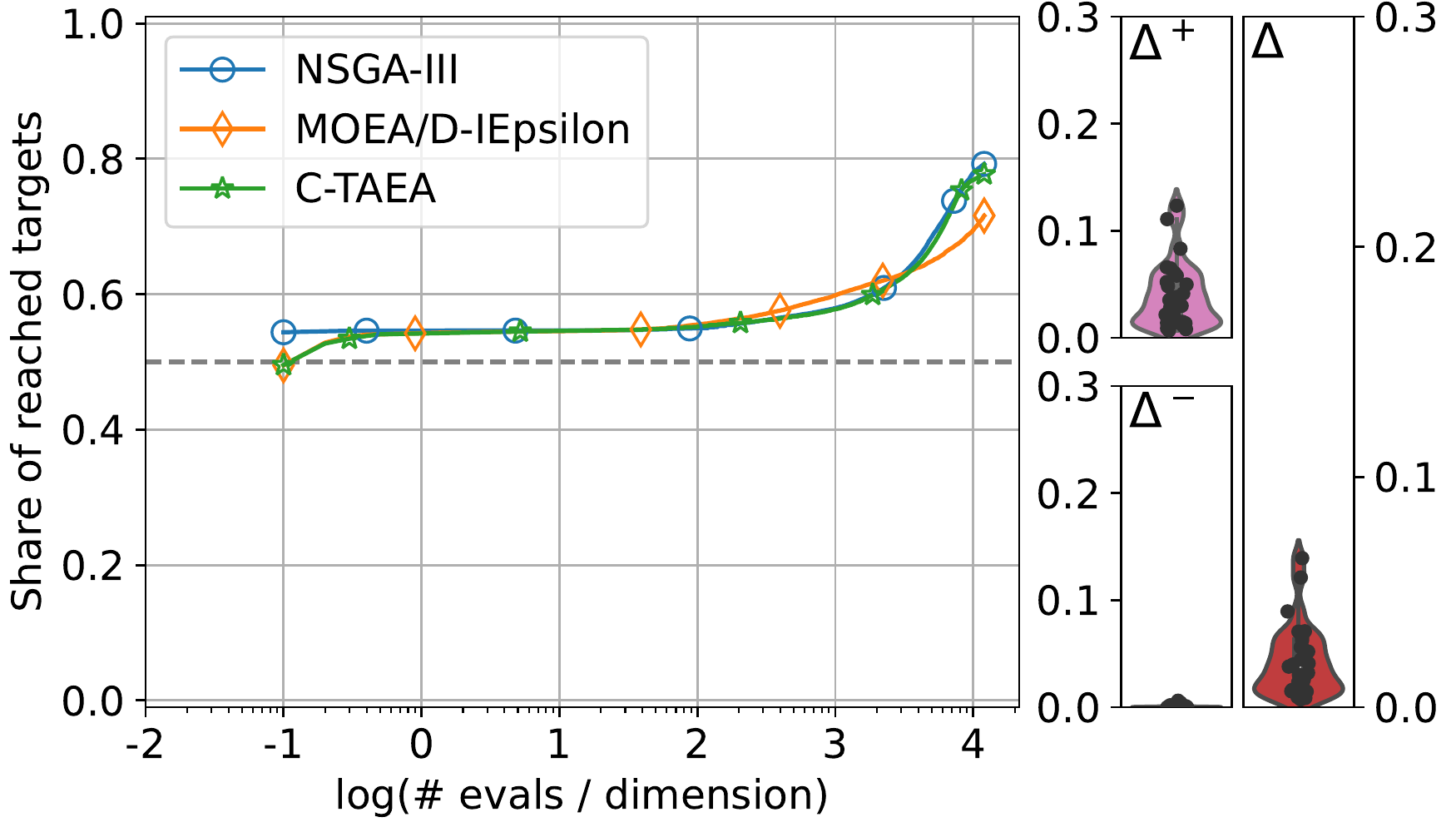}%
    \label{fig:DAS-CMOP_d10}}
    \hfil
    \subfloat[DAS-CMOP ($D=30$)]{\includegraphics[width=0.33\textwidth]{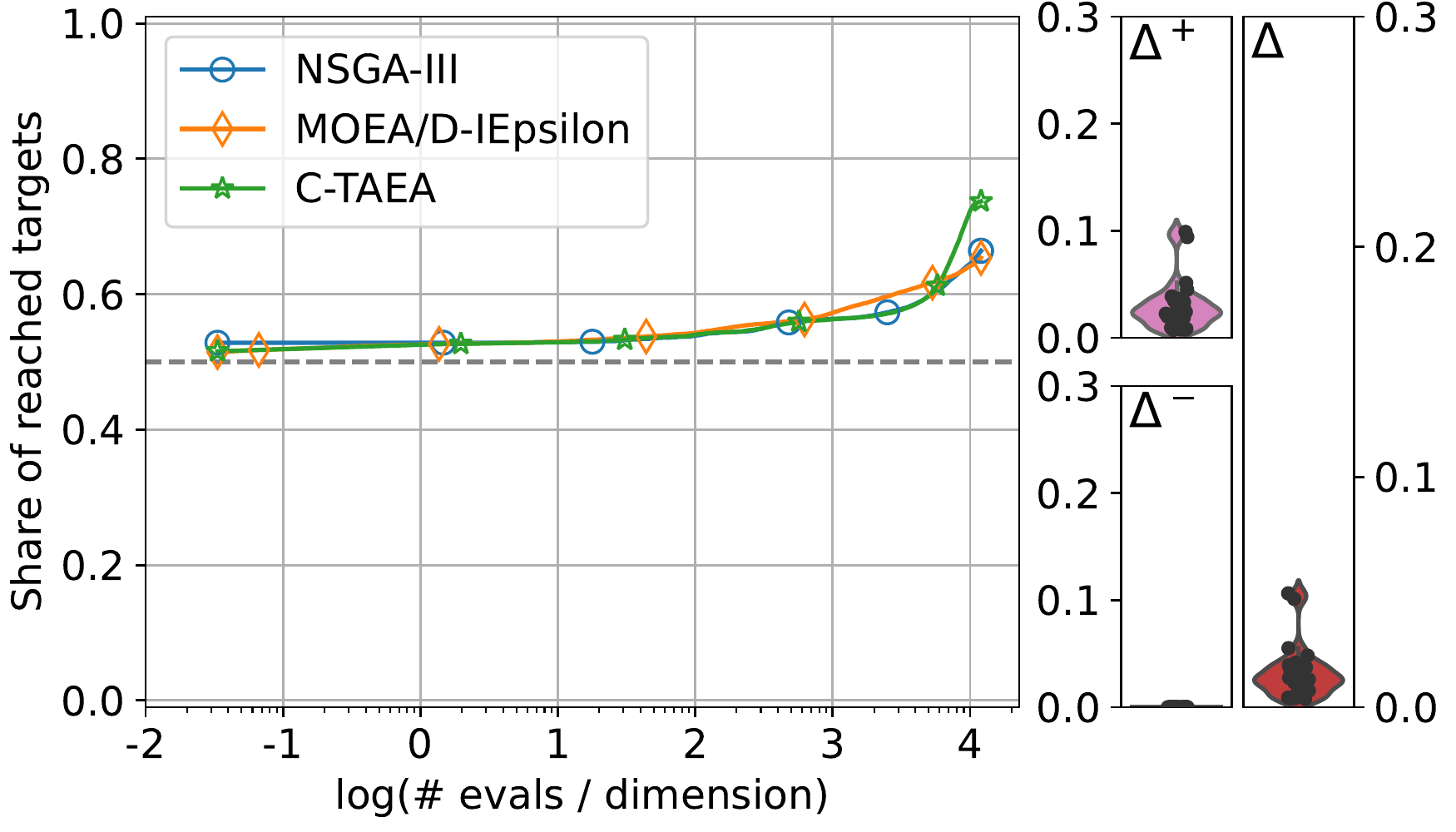}%
    \label{fig:DAS-CMOP_d30}}
    \hfil

    \subfloat[LIR-CMOP ($D=5$)]{\includegraphics[width=0.33\textwidth]{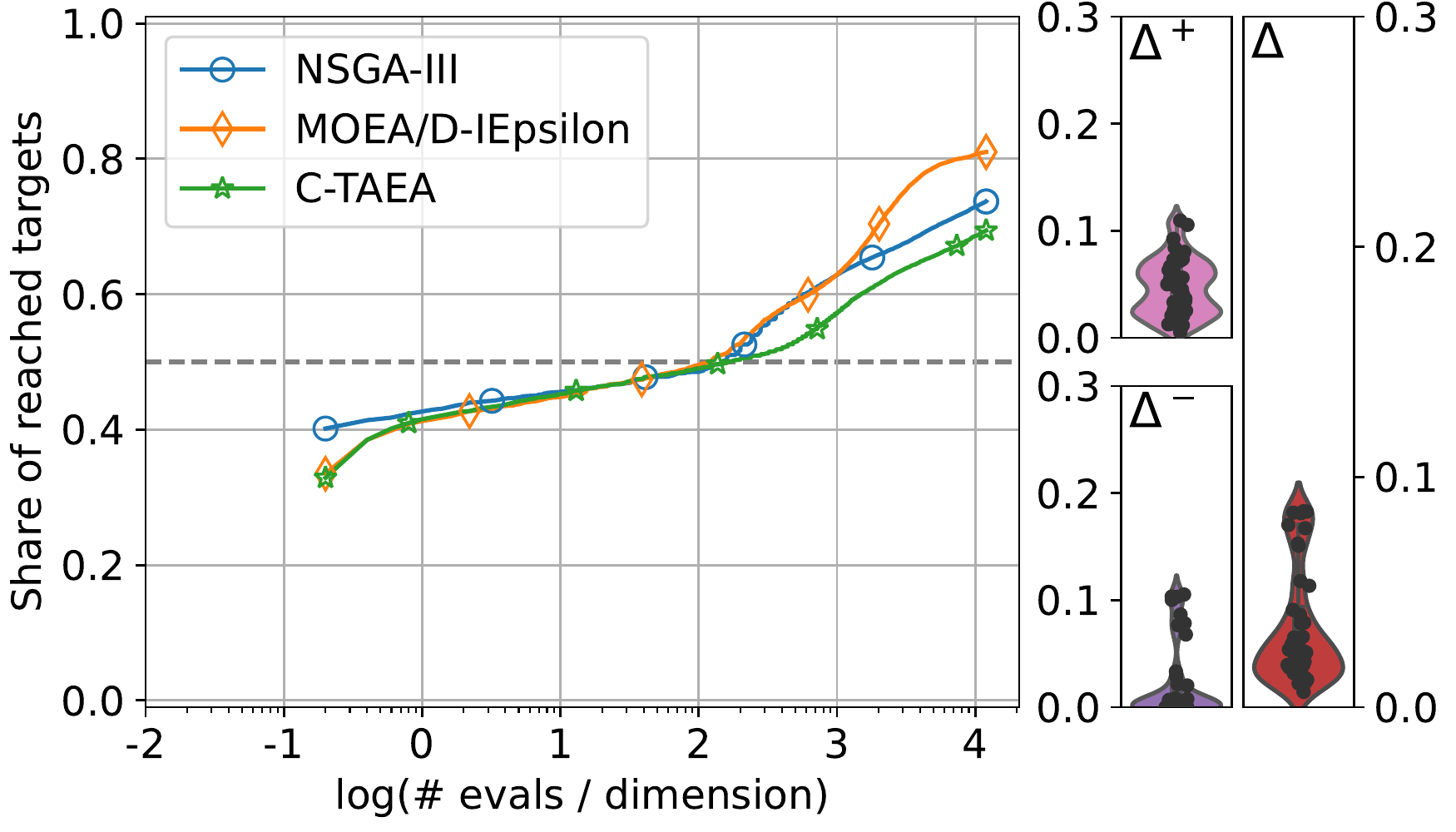}%
    \label{fig:LIR-CMOP_d5}}
    \hfil
    \subfloat[LIR-CMOP ($D=10$)]{\includegraphics[width=0.33\textwidth]{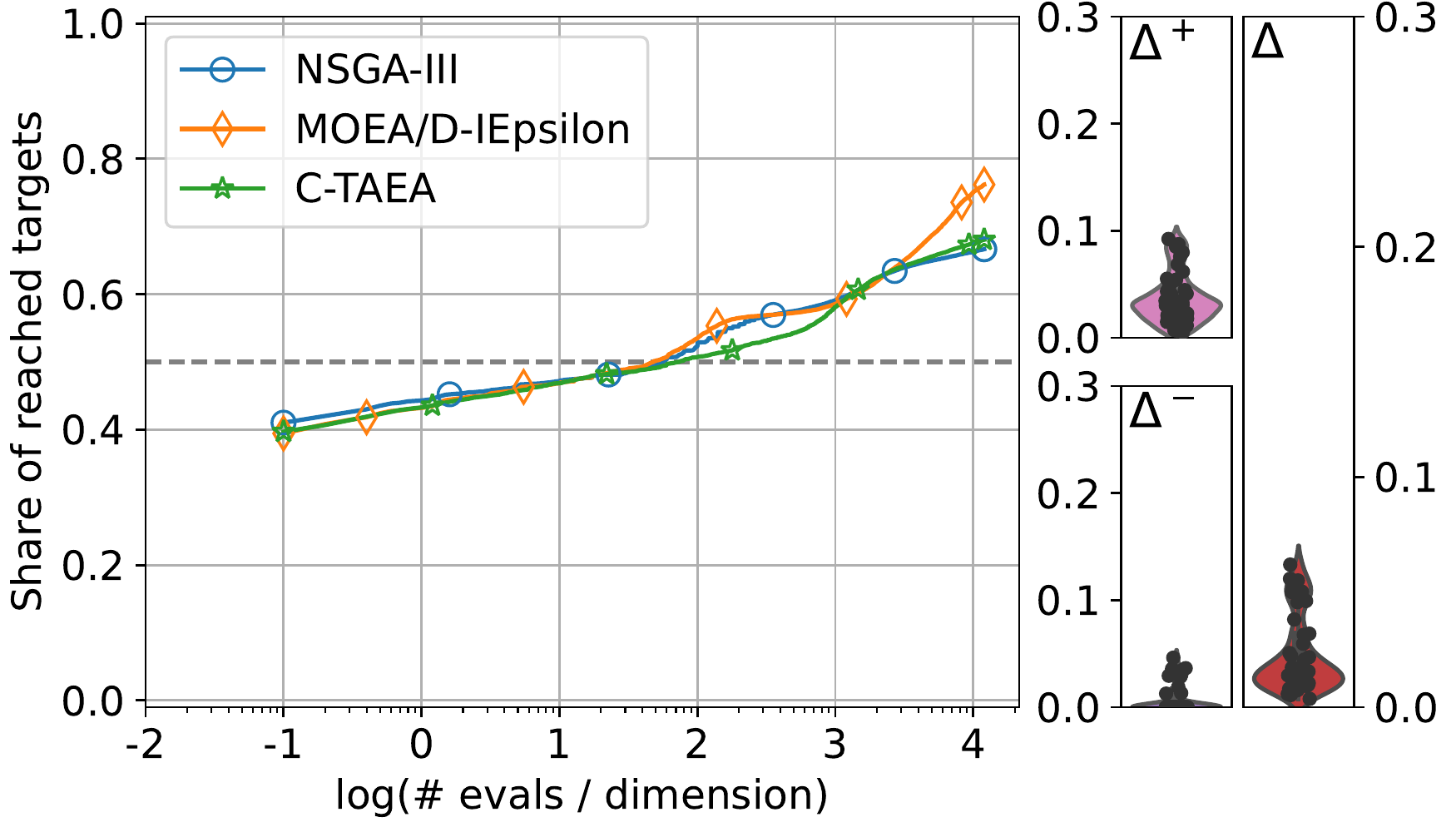}%
    \label{fig:LIR-CMOP_d10}}
    \hfil
    \subfloat[LIR-CMOP ($D=30$)]{\includegraphics[width=0.33\textwidth]{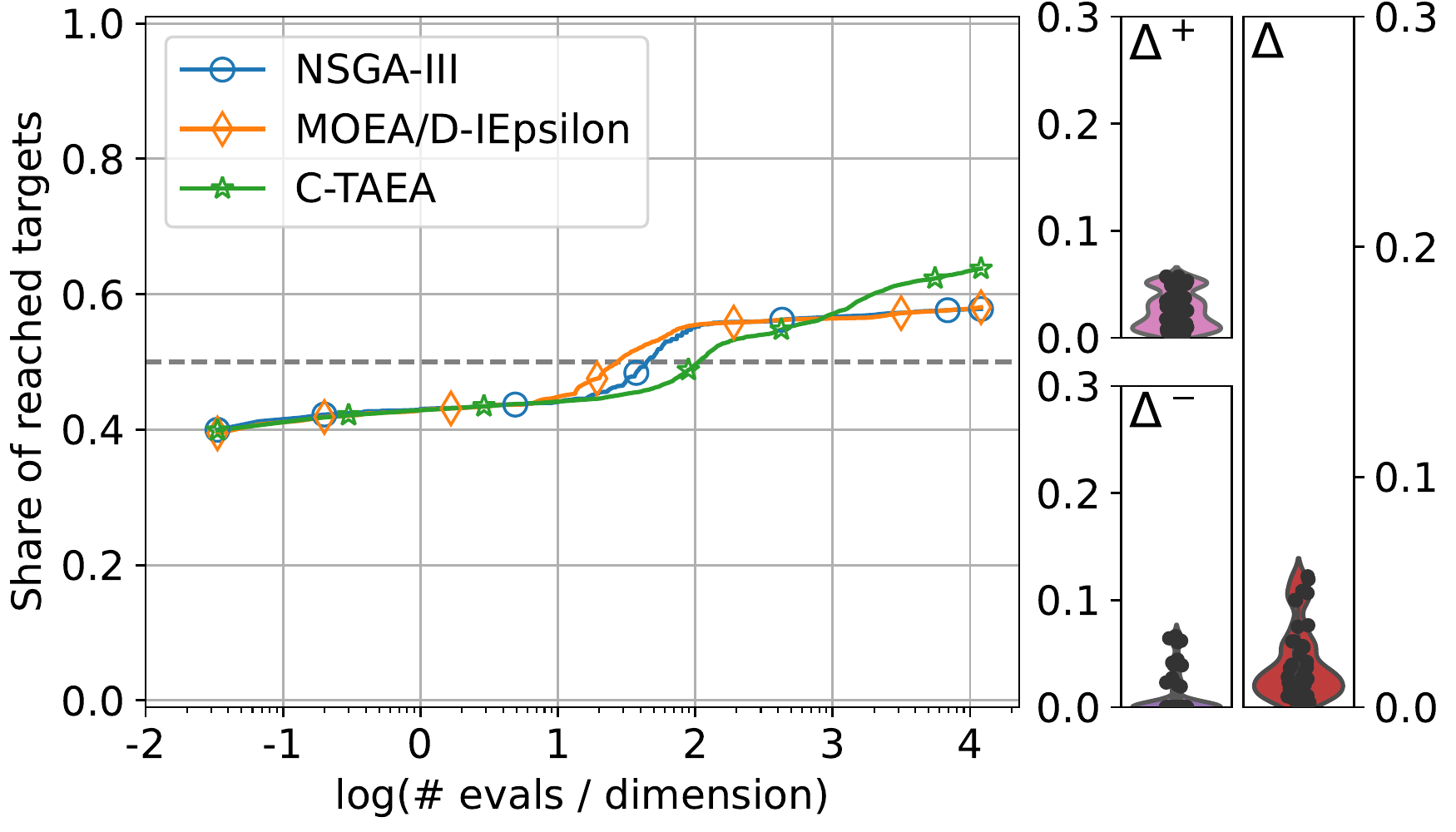}%
    \label{fig:LIR-CMOP_d30}}
    \hfil

    \subfloat[MW ($D=5$)]{\includegraphics[width=0.33\textwidth]{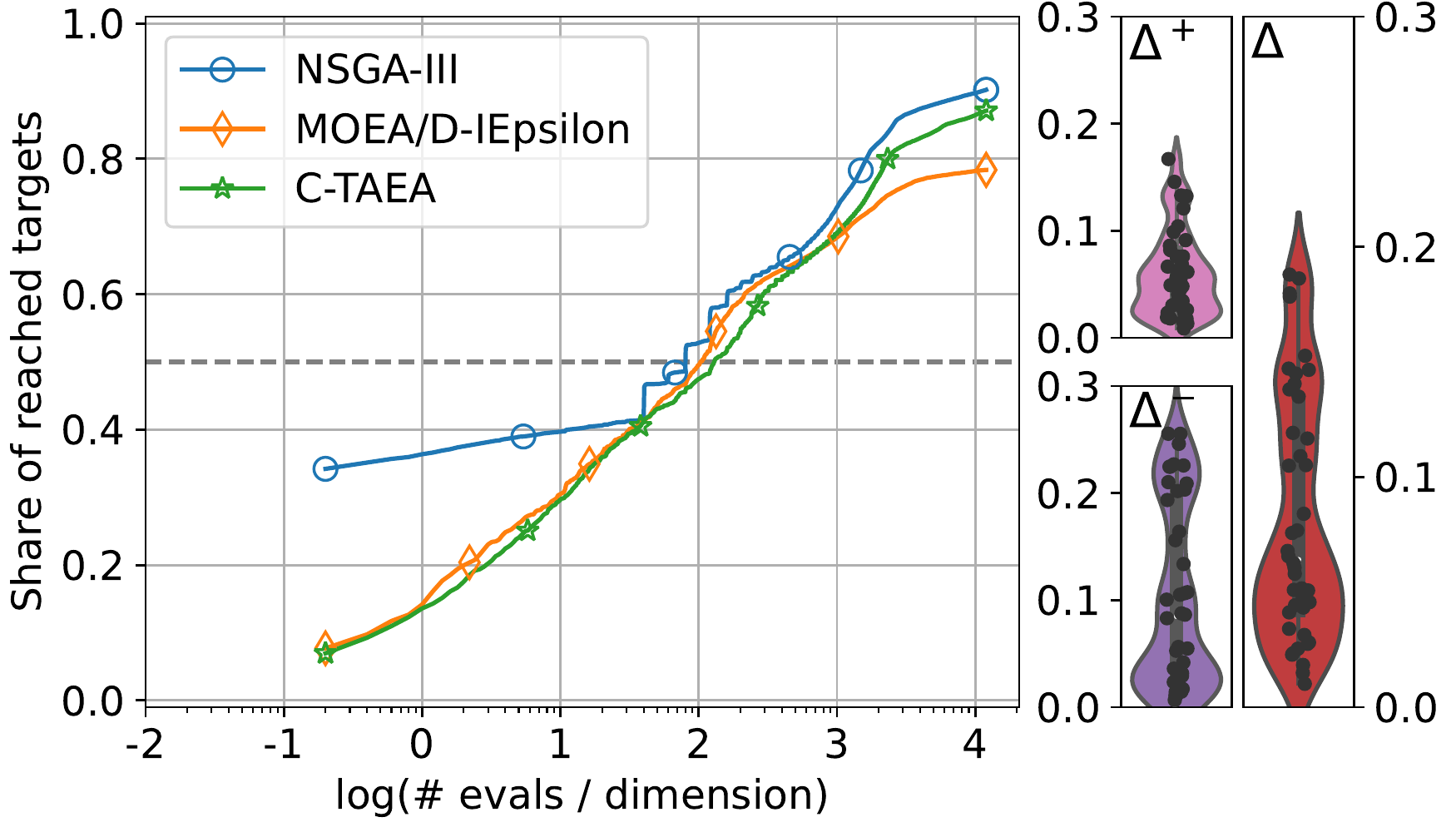}%
    \label{fig:MW_d5}}
    \hfil
    \subfloat[MW ($D=10$)]{\includegraphics[width=0.33\textwidth]{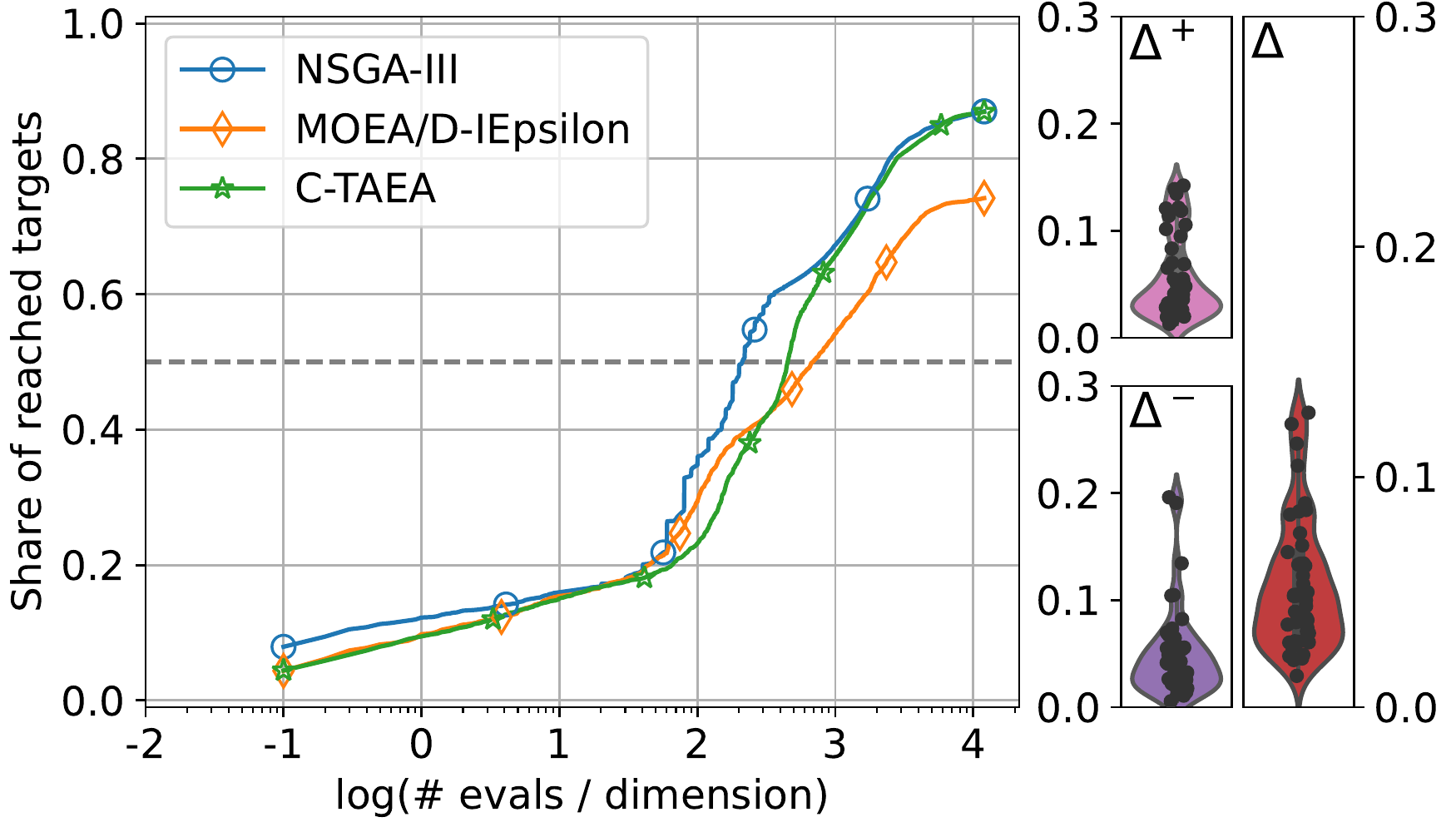}%
    \label{fig:MW_d10}}
    \hfil
    \subfloat[MW ($D=30$)]{\includegraphics[width=0.33\textwidth]{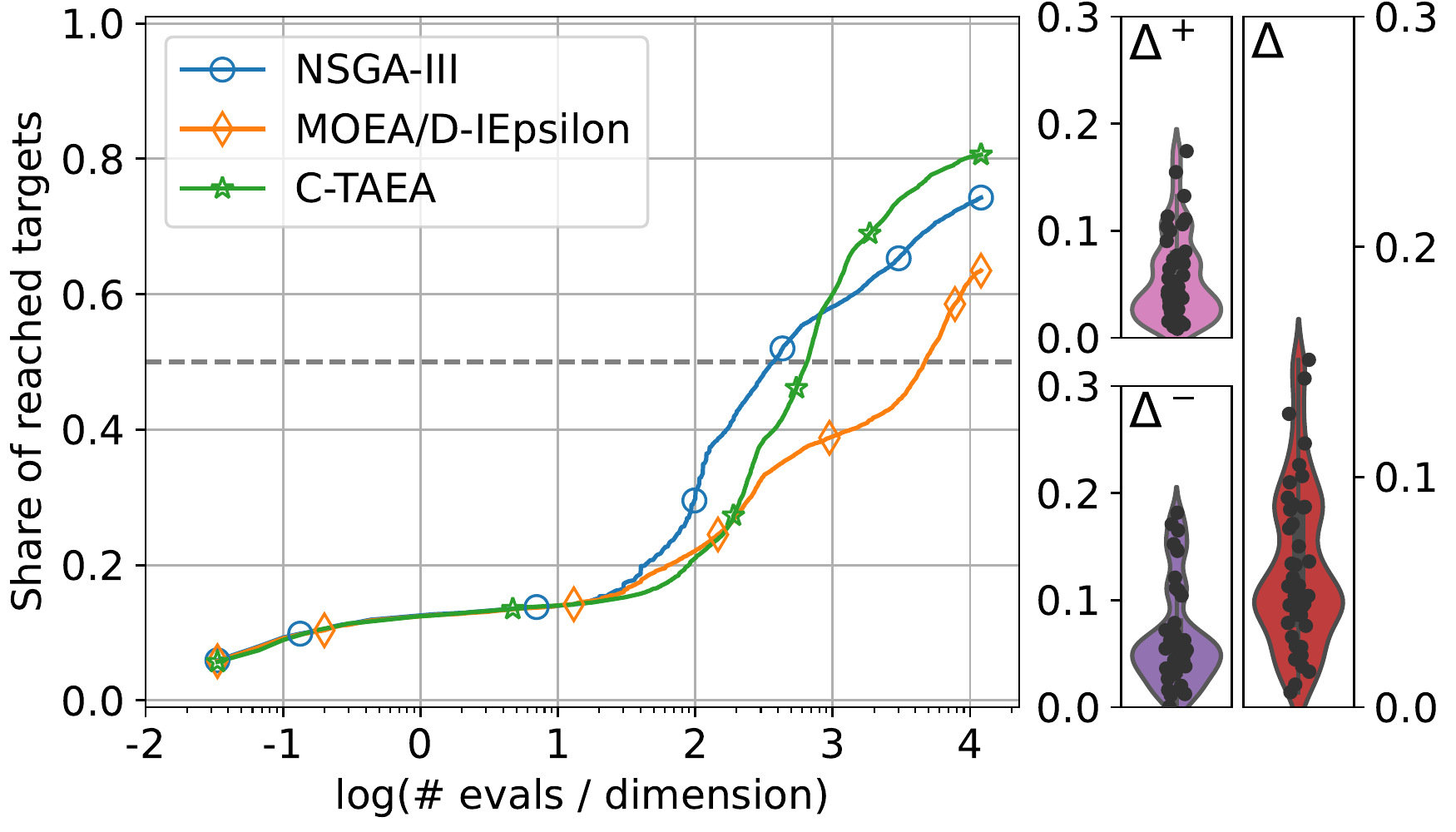}%
    \label{fig:MW_d30}}
    \hfil

    \caption{Results of the three MOEAs on CMOPs from DC-DTLZ, DAS-CMOP, LIR-CMOP, and MW suites. The left plot of each subfigure shows empirical runtime distribution aggregated over all CMOPs in the suite and all targets in dimension 5 (left), 10 (center) and 30 (right). On the right of each subfigure, violin plots depict distributions of $\Delta^+$ (top left), $\Delta^-$ (bottom left), and $\Delta$ (right) values. The larger the diversity, the better.}
    \label{fig:ecdfs_2}
\end{figure*}

The aggregated results over all problems of a suite are shown in Figs.~\ref{fig:ecdfs_1}~and~\ref{fig:ecdfs_2} on the left hand side of each subfigure. For example, Fig.~\ref{fig:CTP_d5} shows the ERDs for the selected MOEAs aggregated over all problems from the CTP suite in 5-$D$. On the right hand side of each subfigure we see violin plots of distributions for $\Delta^+$ (top left), $\Delta^-$ (bottom left), and $\Delta$ (right) values. Each of these values was computed for 30 runs of each pair of algorithms on each problem and is represented in the plot as a black dot. The first column shows the distributions for all the considered CMOPs and the rest of the columns correspond to each suite separately. The $y$-axis depicts the $\Delta^+$, $\Delta^-$, or $\Delta$ value, while the $x$-axis has no specific meaning and is used solely for better visualization. The violin plot (colored area) approximates the probability density function for $\Delta^+$, $\Delta^-$, or $\Delta$ values. For example, Fig.~\ref{fig:MW_d5} shows there are more problem instances in the MW suite with $\Delta \approx 0.05$ than those with $\Delta \approx 0.10$. In addition, there are no problem instances with $\Delta > 0.25$. 

\subsection{Test Suites Evaluation}
As already discussed in Section~\ref{sec:performance_space_comparison}, a well-designed test suite should include a wide variety of problems that can differentiate among MOEAs ($\Delta$). Since we are dealing with constrained problems, we are particularly interested in evaluating the ability of the problems to differentiate among the algorithms with respect to constraint handling ($\Delta^-$).

\begin{itemize}

    \item CTP: As we can see, for all problems the algorithms start in the feasible space (Figs.~\ref{fig:CTP_d5}, \ref{fig:CTP_d10}, and \ref{fig:CTP_d30}). The main difficulty they face is front approximation. This is additionally confirmed by the violin plots showing that $\Delta^-$ equals 0 for all dimensions. 
    
    \item CF: Unlike for CTPs, the MOEAs find no feasible solutions at the very beginning of the evolution process (Figs.~\ref{fig:CF_d5}, \ref{fig:CF_d10}, and \ref{fig:CF_d30}). Interestingly, the difference in algorithm performance originates mainly in constraint satisfaction.

    \item C-DTLZ: From the performance space perspective, this suite is well-designed. The algorithms struggle to find feasible solutions in the initial phase of the evolution process (Figs.~\ref{fig:C-DTLZ_d5}, \ref{fig:C-DTLZ_d10}, and \ref{fig:C-DTLZ_d30}). In addition, the suite can differentiate between algorithms in both constraint satisfaction and front approximation.

    \item NCTP: Although the three MOEAs need a large number of function evaluations to reach a feasible region, their main challenge is front approximation (Figs.~\ref{fig:NCTP_d5}, \ref{fig:NCTP_d10}, and \ref{fig:NCTP_d30}). The vast majority of the difference in algorithm performance also comes from front approximation.

    \item DC-DTLZ: Figs.~\ref{fig:DC-DTLZ_d5}, \ref{fig:DC-DTLZ_d10}, and \ref{fig:DC-DTLZ_d30} show that all the three MOEAs struggle in obtaining feasible solutions, which are discovered only late in the evolution process. Like CFs, the DC-DTLZ suite is especially good at differentiating the constraint satisfaction part of algorithm performance.

    \item DAS-CMOP: As we can see, for all the DAS-CMOPs the NSGA-III algorithm always starts with a feasible solution, while this is not true for other two MOEAs (Figs.~\ref{fig:DAS-CMOP_d5}, \ref{fig:DAS-CMOP_d10}, and \ref{fig:DAS-CMOP_d30}). Nevertheless, feasible solutions are easily discovered by all the algorithms. Moreover, algorithm performance differences are almost exclusively obtained in front approximation, since $\Delta^- \approx 0$ for all problems and MOEAs.

    \item LIR-CMOP: The performance space characteristics of this suite are very similar to those of NCTPs. Although the studied algorithms need some time to find feasible solutions, the main difference in the algorithm performance is contained in the front approximation phase (Figs.~\ref{fig:LIR-CMOP_d5}, \ref{fig:LIR-CMOP_d10}, and \ref{fig:LIR-CMOP_d30}).

    \item MW: From the performance space perspective, MW is one of the most versatile and well-designed artificial test suites found in the literature. It is the best among the studied suites in differentiating the three MOEAs (Figs.~\ref{fig:MW_d5}, \ref{fig:MW_d10}, and \ref{fig:MW_d30}). Moreover, as shown by the violin plots, the algorithm performance is diverse in both constraint satisfaction and front approximation.
    
\end{itemize}

\subsection{Limitations}
We see two potential limitations of the proposed methodology to evaluate the performance space. Firstly, the results can be severely effected by the selection of the algorithms and their budgets, and secondly, the choice of target precision values also has a great impact on the outcome. 

To alleviate the first issue, we selected three different MOEAs equipped with distinct CHTs. Additionally, the number of function evaluations was set large enough to assure convergence thus revealing the deviations between the algorithms. Finally, the logarithmic scale was applied to the budget to not bias the results towards the tail of the convergence graphs where the algorithms have already converged.

On the other hand, we were not able to satisfactorily address the issue of some targets  having greater impact than others. For example, there is no assurance that progressing from target $1 + 10^{-4.2}$ to target $1 + 10^{-4.3}$ is equally important or difficult as progressing from target $1 + 10^{-4.3}$ to target $1 + 10^{-4.4}$ for all the problems and algorithms at hand. Nevertheless, using the target approach and logarithmic scale to define these targets is argued to be much more efficient to compare algorithm performance than just relying on regular convergence graphs~\cite{Hansen2022}.

\section{Conclusions} 
\label{sec:conclusions}

This paper presents a holistic investigation of the existing artificial test CMOPs from a performance space perspective. Firstly, we have proposed a performance assessment methodology capable of simultaneously monitoring both the front approximation and constraint satisfaction. This methodology is an extension of the approach used by the COCO platform for unconstrained bi-objective optimization problems. Next, the resulting performance methodology has been used to analyze and contrast eight artificial test suites. In particular, the test suites have been assessed with respect to the effectiveness of differentiating between three well-known MOEAs. Finally, the paper discusses the advantages and drawbacks of the existing artificial test suites and discloses some limitations of the proposed methodology.

The experimental results show that the CF, DC-DTLZ, and especially MW suites have the greatest potential in differentiating the three MOEAs. They all include multiple CMOPs that can separate the MOEAs in both front approximation and constraint satisfaction. Additionally, our findings indicate that half of the artificial test suites fail to satisfactorily differentiate among the three MOEAs. This suggests that CMOPs from those suites provide limited information for the algorithm designer and are thus of little value for the benchmarking purpose. Finally, we saw that the predominant source of complexity in artificial test CMOPs is the front approximation.

As for the future work, we suggest to extend the proposed methodology to CMOPs with more than three objectives. In particular, since the hypervolume calculation is expensive in high-dimensional objective spaces, one could investigate the effect of using different performance indicators, e.g., inverted generational distance, epsilon indicator, etc. Additionally, the potential of the proposed methodology in studying algorithm behavior while solving real-world problems should also be addressed. Measuring algorithm performance in this case is especially challenging as in a real-world scenario the Pareto front is usually unknown. Finally, the proposed methodology should be tested with additional MOEAs to further support our findings.


%



\ifCLASSOPTIONcaptionsoff
  \newpage
\fi



\bibliographystyle{IEEEtran}
\bibliography{IEEEabrv.bib, bibtex/bib/vodo21a.bib}
%

%

\begin{IEEEbiography}[{\includegraphics[width=1in,height=1.25in,clip,keepaspectratio]{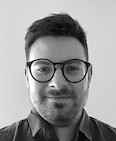}}]{Aljoša Vodopija}
is an assistant researcher at the Department of Intelligent Systems of the Jožef Stefan Institute and a PhD student of Information and Communication Technologies at the Jožef Stefan International Postgraduate School. Within his current doctoral research, he focuses on constrained multiobjective optimization with evolutionary algorithms. In addition, he is involved in developing software solutions for real-world optimization problems. 
\end{IEEEbiography}

\begin{IEEEbiography}[{\includegraphics[width=1in,height=1.25in,clip,keepaspectratio]{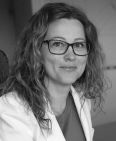}}]{Tea Tušar}
is a research associate at the Department of Intelligent Systems of the Jožef Stefan Institute, and an assistant professor at the Jožef Stefan International Postgraduate School, both in Ljubljana, Slovenia. Her research interests include evolutionary algorithms for singleobjective and multiobjective optimization with emphasis on visualizing and benchmarking their results and applying them to real-world problems. 
\end{IEEEbiography}


\begin{IEEEbiography}[{\includegraphics[width=1in,height=1.25in,clip,keepaspectratio]{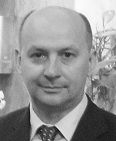}}]{Bogdan Filipič}
is a senior researcher and head of Computational Intelligence Group at the Department of Intelligent Systems of the Jožef Stefan Institute, Ljubljana, Slovenia, and associate professor of Computer Science at the Jožef Stefan International Postgraduate School. He received his PhD degree in Computer Science from the University of Ljubljana. His research interests are in intelligent systems, evolutionary computation and randomized optimization.
\end{IEEEbiography}




\end{document}